\pdfoutput=1

\documentclass[11pt]{article}

\usepackage[]{emnlp2023}

\usepackage{times}
\usepackage{latexsym}

\usepackage[T1]{fontenc}

\usepackage[utf8]{inputenc}

\usepackage{microtype}

\usepackage{inconsolata}
\usepackage{amsmath}
\usepackage{cleveref}
\usepackage{graphicx}

\usepackage[compact]{titlesec}

\usepackage{comment}

\title{When Language Models Fall in Love:\\Animacy Processing in Transformer Language Models\thanks{\ \ This is the camera-ready version of the paper accepted for publication in the Proceedings of EMNLP 2023 (Singapore, December 6-10).}}

\author{Michael Hanna \\
  ILLC \\
  University of Amsterdam \\
  \texttt{m.w.hanna@uva.nl} \\\And
  Yonatan Belinkov \\
  Technion---IIT, Israel\\
  \texttt{belinkov@technion.ac.il} \\\And
  Sandro Pezzelle \\
  ILLC \\
  University of Amsterdam \\
  \texttt{s.pezzelle@uva.nl} \\}

\begin{document}
\maketitle
\begin{abstract}
Animacy\textemdash whether an entity is alive and sentient\textemdash is fundamental to cognitive processing, impacting areas such as memory, vision, and language. However, animacy is not always expressed directly in language: in English it often manifests indirectly, in the form of selectional constraints on verbs and adjectives. 
This poses a potential issue for transformer language models (LMs): they often train only on text, and thus lack access to extralinguistic information from which humans learn about animacy.
We ask: how does this impact LMs' animacy processing---do they still behave as humans do?
We answer this question using open-source LMs. 
Like previous studies, we find that LMs behave much like humans when presented with entities whose animacy is typical. However, we also show that even when presented with stories about atypically animate entities, such as \textit{a peanut in love}, LMs adapt: they treat these entities as animate, though they do not adapt as well as humans. Even when the context indicating atypical animacy is very short, LMs pick up on subtle clues and change their behavior. We conclude that despite the limited signal through which LMs can learn about animacy, they are indeed sensitive to the relevant lexical semantic nuances available in English.
\end{abstract}

\section{Introduction}
Animacy plays a significant role in cognitive processing, as evidenced by the fact that animate entities are easier to remember and prioritized in visual processing \citep{nairne2013adaptive,new2007category,bugaiska2019animacy}. It is so important that even young children can distinguish animate and inanimate entities \citep{rakison2021developmental}, and these are processed in distinct domain-specific brain regions \citep{caramazza1998domain}.

Animacy distinctions also manifest in language; however these distinctions may appear indirectly. While some languages explicitly mark animacy, animacy distinctions in English often take the form of selectional constraints that limit the use of certain verbs or adjectives with in/animate entities. For example, only animate entities can \textit{walk} or \textit{think}. So, while animacy is a rich distinction at the cognitive level, at the linguistic level, its signal can be muted.


Today's pre-trained transformer language models (LMs), however, are trained only on linguistic input. If they are to learn to process animacy, they must thus do so only from its downstream effects in text, unlike humans, who use visual and physical stimuli. We therefore ask: do such LMs respond to animacy in language as humans do?

We answer this by treating LMs as psycholinguistic test subjects, probing how they react to violations of animacy-related selectional constraints. Like prior work \citep{warstadt-etal-2020-blimp-benchmark, kauf2022event}, we first study LMs' responses in scenarios involving typical animacy. In such situations, animacy is a simple mapping between an object (e.g. \textit{a peanut}) and its usual animacy (inanimate). We find that like humans, LMs generally prefer sentences that respect animacy-related selectional constraints, assigning higher probabilities to such sentences. 

Unlike prior work, we also study \textit{atypical} animacy \citep{coll-ardanuy-etal-2020-living}, where a typically inanimate object becomes animate. We draw on \citet{Nieuwland2006WhenPF}, which measured human N400 responses in scenarios with atypically animate entities like \textit{a peanut in love}. We compare LM surprisal to human N400 brain responses and find that like humans, LMs are initially surprised to encounter entities like \textit{a peanut in love}, but quickly adapt, becoming less surprised. Stronger LMs are more able to replicate the large magnitude of human N400 reduction. 

Given LMs' success at adapting to atypical animacy with a long context, we test them on short sentences about atypically animate entities, and measure the extent to which their outputs reflect this atypical animacy. We find that even with limited context, LMs adapt their output distribution, treating the entity as animate. We conclude that, despite training without the modalities that humans use to learn about animacy, LMs respond to shifting animacy in a surprisingly human-like way. Code for our experiments is available at \url{https://github.com/hannamw/lms-in-love}.

\section{Related Work}
\subsection{Animacy in Language} 

Animacy in cognition is often framed as a gradient phenomenon \citep{deSwart2018animacy}. In language, this often simplifies to a tripartite hierarchy (humans > animals > objects) or a binary (humans \& animals >\ objects); entities are distinguished synactically or morphologically by their position therein \citep{comrie1989language}. 

Animacy exists at both the type level (e.g.\ dogs > rocks) and the token level (e.g.\ a specific rock in a story might be animate, though rocks are typically not). Moreover, linguistic animacy is based not only on biology, but also on the speaker's closeness and empathy with the entity in question \citep{kuno1977empathy}; thus a speaker might treat their dog as more animate than an unknown dog.

The precise effects of animacy in language vary cross-linguistically, from explicit animacy marking to more indirect effects as in English. The latter include not only strict animacy-based selectional constraints \citep{caplan1994selectional}, but also more subtle grammatical influences \citep{rosenbach2008animacy,bresnan2008animacy}. For example, animate entities are more often mentioned first in a sentence, even if doing so produces less common constructions, such as the passive \citep{ferreira1994passive}.

Here, we focus on the human / inanimate object dichotomy, and the animacy-based selectional constraints thereby imposed; this strong contrast should produce easier-to-measure effects in LMs.

\subsection{LMs as Test Subjects} \label{sec:lms-test-subjects} \label{sec:n400-surprisal}
We study the behavior of LMs by treating them as psycholinguistic test subjects, a popular approach. One such line of work analyzes LMs by using the probability they assign to a sentence as a proxy for acceptability judgments. Generally, such studies provide pairs of sentences, one acceptable and one not; LMs must assign the more plausible sentence a higher probability. This method has been used to study LMs' processing of negation, subject-verb agreement, and more \citep{ettinger-2020-bert,linzen-etal-2016-assessing,warstadt-etal-2020-blimp-benchmark,sinclair-etal-2022-structural}.

Other work compares LMs to humans by using surprisal---the negative log probability of a string---to estimate measures of cognitive effort during text processing. LM surprisal is versatile, and well-correlated with reading times, eye-tracking fixations, and EEG responses \citep{smith2013predictability,aurnhammer2018comparing,michaelov-bergen-2020-well}; moreover, surprisal from stronger LMs provides better predictive power \citep{goodkind-bicknell-2018-predictive,wilcox2020predictive}.

We use LM surprisal to predict the N400 brain response, which is elevated at semantically unusual content, like animacy-related selectional constraint violations. Studies have found that a word's surprisal correlates with human N400 response thereto \citep{frank-etal-2013-word,FRANK20151,michaelov2022n400}; transformer LMs are the state of the art for this \citep{merkx-frank-2021-human,michaelov2021different}. 

\subsection{Animacy Detection} 
We note that our interest in the processing of atypical animacy parallels similar developments in the NLP task of \emph{animacy detection}: determining whether a given entity is animate. While many animacy detection studies originally considered only typical animacy \citep{orasan2007animacy,bowman-chopra-2012-automatic}, later animacy detection work has recognized that entities' animacy may not always be typical, and may change over the course of a narrative \citep{karsdorp2015animacy,jahan-etal-2018-new}. Particularly relevant for the present study, \citet{coll-ardanuy-etal-2020-living} combine LMs and atypical animacy by using BERT for atypical animacy detection. Although this approach uses LMs as a tool to label animacy, rather than studying how LMs process animacy, we share their interest in the atypical edge cases of animacy.

\subsection{Animacy in LMs}
How neural models capture animacy has long interested cognitive scientists; \citet{elman1990finding} trained a simple neural LM on artificial language data, and found that its representations of animate and inanimate entities formed distinct clusters. More recent work has assessed the animacy-processing capabilities of modern LMs, mostly focusing on typical animacy. Animacy is one area tested by BLiMP \citep{warstadt-etal-2020-blimp-benchmark}, which we revisit in \Cref{sec:blimp}. \citet{kauf2022event} investigate animacy as part of LMs' generalized event knowledge; they also find that LMs are sensitive to (typical) animacy as it pertains to selectional constraints.

We move beyond typical animacy to atypical animacy by using LM surprisal to replicate \citeposs{Nieuwland2006WhenPF} studies on human N400 response to atypical animacy.
Contemporaneous work \citep{michaelov2023peanuts} replicates one of these experiments in the original Dutch. In contrast, we replicate all experiments from \citeauthor{Nieuwland2006WhenPF} (and \citet{boudewyn2019adaptation}). These highlight situations in which models can capture general trends, but fail to capture low-level nuances. Moreover, by studying a diverse set of English LMs, we can identify how LMs' strength affects their predictive power.

\section{Models}\label{sec:models}
We experiment with these models: GPT-2 small, medium, large, and XL \citep{radford2019language}; OPT 2.7B, 6.7B, and 13B \citep{zhang2022OPT}; and LLaMA 7B, 13B and 30B \citep{touvron2023llama}.\footnote{The names of OPT and LLaMA models indicate (approximate) parameter counts; the GPT-2 models have 117M, 345M, 762M, and 1.5B parameters respectively.} We use autoregressive LMs, as we need to compute probabilities for whole sentences. Moreover, we choose open-source models, to make our work replicable. We provide implementation details in \Cref{app:implementation-details}.

\section{Typical Animacy}\label{sec:blimp}
We test models' responses to animacy in situations where the animacy of a given token, or instance of entity, aligns with the animacy of its type more generally (e.g. cats are animate; rocks are not).

\paragraph{Experiment} We test the models in \Cref{sec:models} on the \textit{animate-transitive} and \textit{animate-passive} datasets of the BLiMP benchmark \citep{warstadt-etal-2020-blimp-benchmark}. Each dataset contains 1,000 minimal pairs of synthetic English sentences that differ only by one or two words (\Cref{tab:blimp-examples}). By construction, one sentence respects animacy constraints; the other violates them. We evaluate models on these datasets by computing the probability it assigns to each sentence of each minimal pair. A model gets an example correct if it assigns higher probability to the sentence that respects the animacy constraint. We compute model accuracy over each dataset.

\begin{table}
    \centering
    \begin{tabular}{c|c|c}
          &  Acc? & Sentence\\
         \hline
         \textbf{T} & Yes & \textbf{Naomi} had cleaned a fork.\\ 
         \textbf{T} & No & \textbf{That book} had cleaned a fork.\\ 
         \hline
         \textbf{P} & Yes & Lisa was kissed by the \textbf{boys}.\\
         \textbf{P} & No & Lisa was kissed by the \textbf{blouses}.\\
    \end{tabular}
    \caption{BLiMP examples: we provide one example each from the \textbf{T}ransitive and \textbf{P}assive datasets. Each is a minimal pair of sentences: one \textbf{Acc}eptable and one not.} 
    \label{tab:blimp-examples}
\end{table}

\paragraph{Results}

\begin{figure}
    \includegraphics[width=\linewidth]{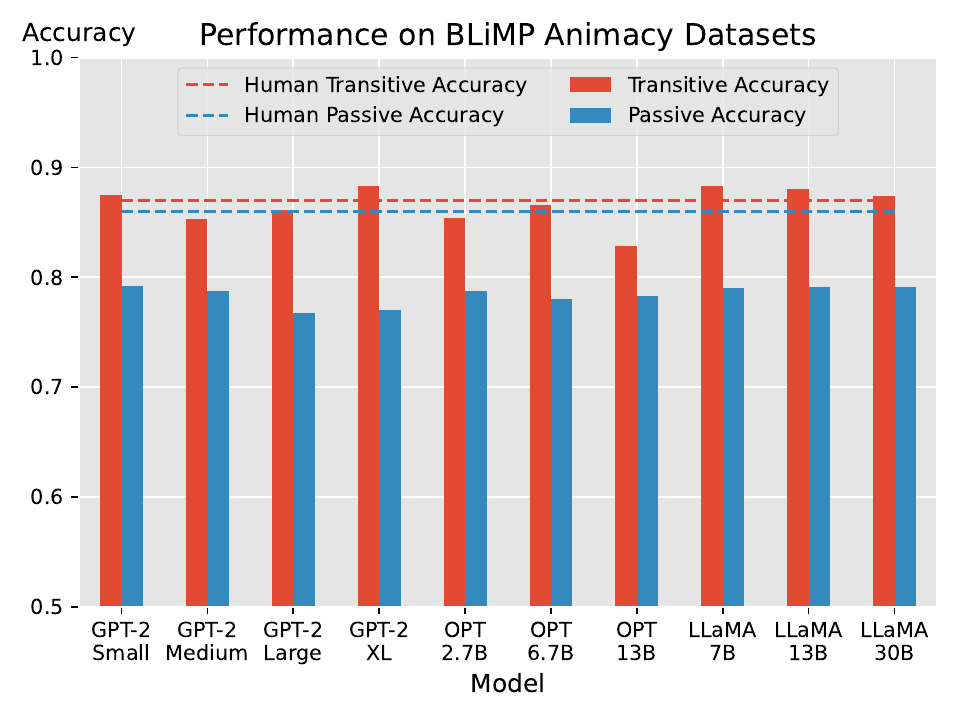}
    \caption{Model accuracy on BLiMP. Models match humans in the transitive, but not passive setting.}
    \label{fig:blimp-bar-chart}
\end{figure}

\Cref{fig:blimp-bar-chart} displays results for each model. It also includes human baselines, reported directly from \citet{warstadt-etal-2020-blimp-benchmark}, which indicate the proportion of examples where annotators preferred the acceptable sentence of the given minimal pair. Random performance for both datasets is 50\%.

Models attain high performance in both scenarios. On transitive examples, they reach over 80\% accuracy; some models prefer the sentence that respects animacy constraints more often than humans do (>87\%). In the passive scenario, the gap between models (80\%) and humans (86\%) is wider.

This difference between the transitive and passive cases may be due more to setup differences than distinct animacy processing in the two scenarios. In the passive case, the target word is always in the last position, so model performance is determined only by the target's probability. In contrast, the target word is not the final token in the transitive case, so model success is determined by the probability of a longer string.

\paragraph{Discussion}
Our results indicate that models respect animacy constraints in typical scenarios: they match human performance on the transitive dataset, and are close behind on the passive. However, this test cannot distinguish between a model that truly understands animacy, and one that just associates words (types) with other words that reflect that word's typical animacy. For example, the model might simply associate a word like ``shoe'' with verbs that take inanimate objects, without understanding that the inanimacy of an individual shoe is what prohibits its use with animate-selecting verbs.

To solve this problem, our analysis must move beyond type-level animacy, and test models' processing of animacy at the token level. We thus test models' responses to entities whose token-level animacy is atypical, distinct from their usual type-level animacy. If models process these entities according to their type-level animacy, their understanding of animacy is rather shallow. In contrast, models that process entities according to their token-level animacy may better understand animacy in full.

\section{Atypical Animacy}
In this section, we attempt to determine if LMs can capture animacy not only at the type-level, but also at the token-level. We do so by comparing model and human responses in cases of atypical animacy, where entities' canonical type-level animacy and their actual token-level animacy differ.

For human data, we turn to two similar studies---\citet{Nieuwland2006WhenPF} and \citet{boudewyn2019adaptation}---that relied on the N400, a brain response measured via EEG that is elevated when processing semantically anomalous input. Both studies measured participants' N400 responses while they read stories where a typically inanimate entity acted as animate (\Cref{fig:nvb_exp1}), similar in tone and content to a cartoon, or fairy-tale. Both found that while participants were initially surprised by the atypically animate entity, they quickly adapted, yielding low N400 responses to the entity.

We ask if the same is true of pre-trained LMs: can they adapt to entities that are animate at the token-level, despite being typically inanimate? Or is their processing of animacy limited to a simple type-level understanding? To answer this question, we replicate these studies with pre-trained LMs, using their surprisal to model N400 responses.

We replicate three experiments: \citeauthor{Nieuwland2006WhenPF}'s \textbf{repetition experiment}; their \textbf{context experiment}; and \citeauthor{boudewyn2019adaptation}'s \textbf{adaptation experiment}.\footnote{In \Cref{app:bdw_exp2}, we replicate \citeauthor{boudewyn2019adaptation}'s English version of \citeauthor{Nieuwland2006WhenPF}'s context experiment with LMs; our results are identical to those of \Cref{sec:nvb_exp2}.} For each, we first explain the original study. Then, we explain how we adapt the experiment for LMs. Finally, we report our results and compare them the original study's results.

\subsection{Repetition Experiment}\label{sec:nvb_exp1}
\begin{figure}
    \fbox{
  \footnotesize   \parbox{0.46\textwidth}{
    A nurse was talking to the \textbf{sailor/oar [1]} who had been in a violent boating accident. The sailor/oar cried for a long time over the storm that had raged over the lake for hours. The nurse consoled the \textbf{sailor/oar [3]}, saying that he would soon be well again. The sailor/oar complained of a bad headache that would not go away. The nurse gave the \textbf{sailor/oar [5]} a large dose of aspirin. The sailor/oar thanked her and fell asleep.
    }}
    \caption{Story from \citeauthor{Nieuwland2006WhenPF}, repetition experiment (translated and edited). Times when N400 responses were recorded are numbered, in bold.}
    \label{fig:nvb_exp1}
\end{figure}

\paragraph{Original Study} In \citeauthor{Nieuwland2006WhenPF}'s first experiment, participants listened to Dutch stories that contained either a typical, animate entity or an inanimate entity behaving as if it were animate (\Cref{fig:nvb_exp1}). Participants' N400 responses were measured at the 1st, 3rd, and 5th mentions of the entity (in \Cref{fig:nvb_exp1}, either \textit{oar} or \textit{sailor}, in bold). 

\citeauthor{Nieuwland2006WhenPF} found that participants had a moderate N400 response to the first mention of a typically animate entity, and a low response on subsequent mentions. In contrast, participants initially had a high N400 response to the atypically animate entity. However, by the 3rd and 5th mentions thereof, their N400 responses were so low as to be statistically indistinguishable from the responses to mentions of the animate entity in the same contexts. Thus, while humans were initially surprised by the atypically animate entity, they quickly adapted to the situation, and found it no more surprising than typically animate entities.

\paragraph{Our Experiment} 
We model N400 responses with LM surprisal, as discussed in \Cref{sec:n400-surprisal}. For each of the 60 examples, we measure the surprisal of the animate and inanimate entity given the context at each timestep. For example, to model the inanimate N400 response at T1 in the example from \Cref{fig:nvb_exp1} given a model $p_\theta$, we compute $-\log_2 p_\theta(\text{\textit{oar}}|\text{\textit{A nurse was talking to the}})$. Then, we compute the mean surprisal of examples containing animate and inanimate entities separately.

\begin{figure}
    \includegraphics[width=\linewidth]{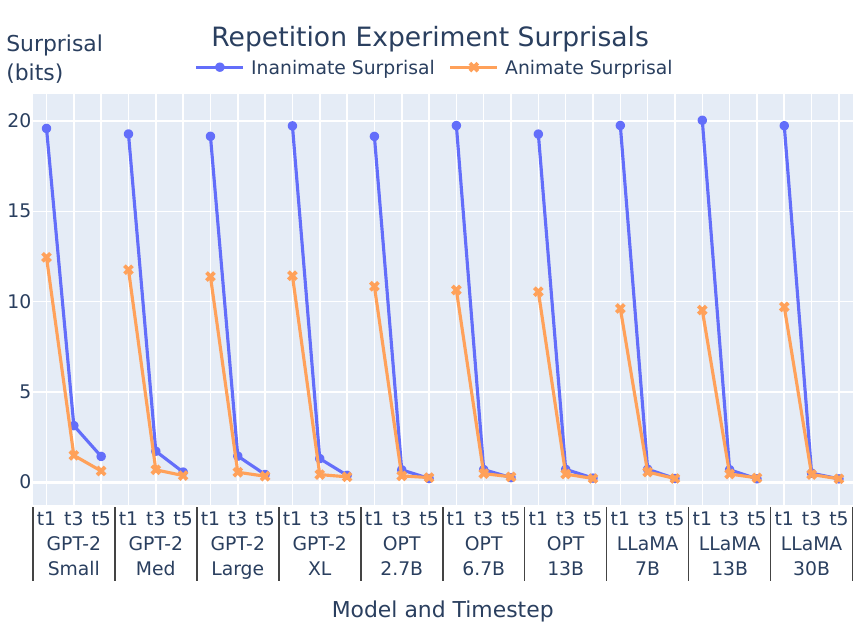}
    \caption{Mean repetition experiment surprisal. Inanimate surprisal is initially higher, but both surprisals decrease rapidly after T1, becoming near identical.}
    \label{fig:nvb_exp1_bar}
\end{figure}

Since the original stimuli are in Dutch, we translate them to English, to make them compatible with the English LMs.\footnote{We perform these experiments in Dutch in \Cref{app:dutch-data}. Dutch results are comparable to English results.} We do so using DeepL;\footnote{\url{https://www.deepl.com/translator}} translations were checked by a native Dutch speaker. We then manually post-edited each stimulus to ensure it matched the cartoon-like tone and content of the original, and contained inanimate characters that violated typical animacy constraints in the 1st, 3rd, and 5th sentences of the stories.\footnote{We also edited stories for fluency, and to convert Dutch cultural references to Anglophone counterparts. The translated English stimuli can be found at \url{https://github.com/hannamw/lms-in-love}} Because we preserve the relevant aspects of the stimuli we expect the trends in N400 responses to be the same.

\paragraph{Results} All models capture broad trends in human N400 responses well (\Cref{fig:nvb_exp1_bar}). At T1, models are very surprised by the inanimate entity, and only moderately surprised by the animate entity. At later timesteps, however, both entities' surprisals drop precipitously, to similar levels: models adapt to both entities quickly, just like humans do.

\begin{figure}
    \includegraphics[width=\linewidth]{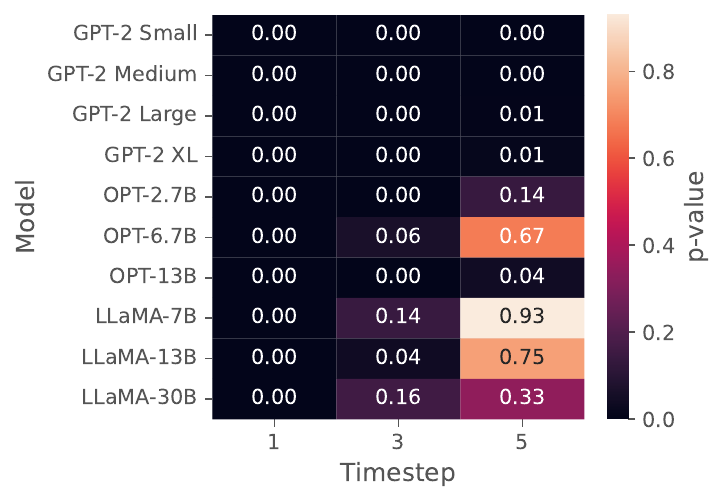}
    \caption{Stat. significance of the difference between animate and inanimate surprisal, by model and timestep}
    \label{fig:animacy_tests}
\end{figure}

Still, the raw results do not prove that models are adapting to the extent that humans are. Since \citeauthor{Nieuwland2006WhenPF} found that human N400 responses in the two conditions were statistically indistinguishable by T3, we test if the same is true for model surprisal. We use the Wilcoxon signed-rank test for non-normally distributed data \cite{wilcoxon1945ranking} to determine if model surprisals for animate and inanimate entities are distinct at each timestep. We find (\Cref{fig:animacy_tests}) that like humans, LMs have a statistically significant difference between animate and inanimate surprisals at T1. However, while there was no difference in humans at T3, there are differences ($p<0.01$) in most models; only the largest exhibit none. At T5, differences disappear in yet more large models. While models can generally approximate trends in human N400 responses to atypical animacy, only the largest and most powerful fully replicate human adaptation.

Overall, pre-trained LMs seem able to mimic human-like adaptation to atypically animate entities. It is tempting to conclude that they have a human-like understanding of animacy, that works at the token rather than the type level. However, it is equally possible that their decreased surprisal is due to repetition, rather than a deeper understanding of animacy. Transformer LMs even have a low-level emergent structure, induction heads, dedicated to such copy-pasting \citep{olsson2022context}.

Fortunately, \citeauthor{Nieuwland2006WhenPF} shared this concern: humans might generate lower N400 responses only because they had seen the atypically animate token before. Thus, we also replicate their context experiment, which avoids this issue.

\begin{figure}
    \fbox{\footnotesize\parbox{.465\textwidth}{
    A girl sat next to a diamond who was always doing strange things. The diamond told her that he liked to eat erasers. The girl ignored the diamond and his stories. Then the diamond said he also liked to sing songs. The diamond was quite \textbf{foolish/valuable} but secretly also very funny. That's why she always sat next to him.
    }}
    \caption{Story from \citeauthor{Nieuwland2006WhenPF}, context experiment (translated and edited). N400 responses were recorded at the words in bold.}
    \label{fig:nvb_exp2}
\end{figure}

\subsection{Context Experiment} \label{sec:nvb_exp2}
\citeauthor{Nieuwland2006WhenPF}'s context experiment showed that participants' low N400 responses did not stem from lexical repetition. 

\paragraph{Original Study} As in the prior experiment, participants read 60 Dutch stories containing an atypically animate entity; at the end of each story, the entity was described using an adjective that was either context-appropriate (and generally used for animate entities) or context-inappropriate (but typical for the inanimate entity; \Cref{fig:nvb_exp2}). The N400 response was measured at the adjective at the end of the story (\textit{foolish} or \textit{valuable}). N400 responses for the context-appropriate animate adjective were far lower than those to the entity-appropriate inanimate adjective, showing that the first experiment's effects were not caused by lexical repetition.

\paragraph{Our Experiment} We calculate the surprisal at the animate and inanimate adjective for each of the 60 stories. We also compute baseline surprisals, the surprisal of the inanimate adjective without the entire story context, to show they are indeed high: e.g. $-\log_2p_\theta(\text{\textit{foolish}}|\text{\textit{The diamond was quite}})$.

\begin{figure}
    \includegraphics[width=\linewidth]{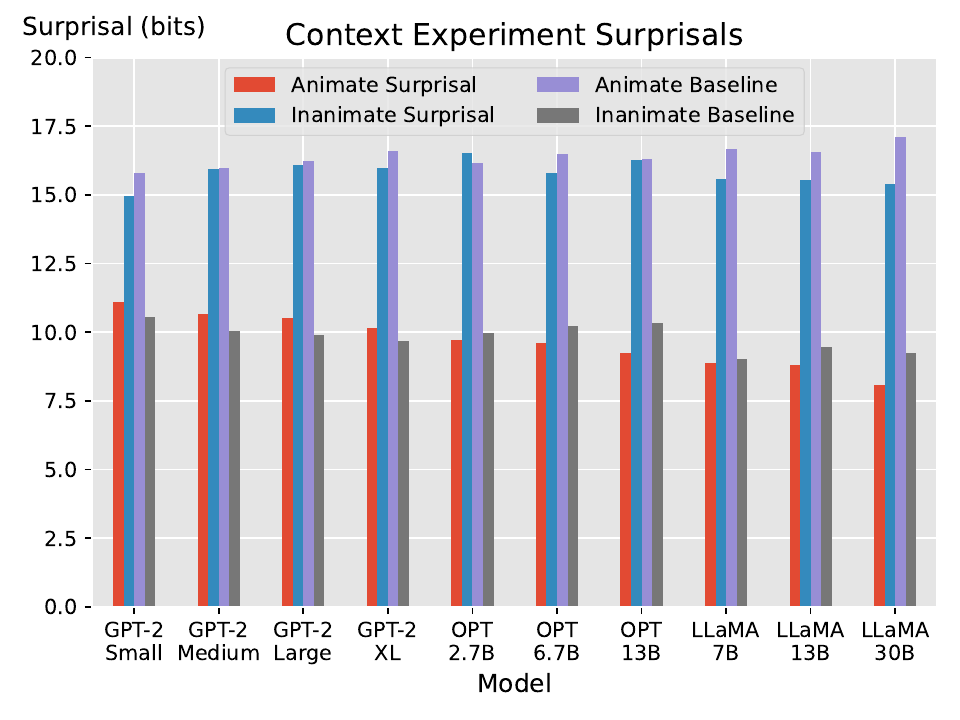}
    \caption{Context experiment surprisals. With context, the animate adjective is much less surprising; in the contextless baseline condition, this is reversed.}
    \label{fig:nvb_exp2_bar}
    \vspace{-5pt}
\end{figure}

\paragraph{Results} For all models, mean surprisal of the animate adjective is much lower than that of the inanimate adjective (\Cref{fig:nvb_exp2_bar}). This is significant in all cases ($p < 0.01$; Wilcoxon signed-rank test). Moreover, the animate adjective is assigned higher probability in almost all cases---over 90\% for large models. This mirrors the human trend: the N400 response to the context-appropriate animate adjective was much lower than the response to the entity-appropriate inanimate adjective. In the contextless baseline situation, the inanimate adjective receives a much lower surprisal than the animate adjective. Like humans, models use context and overcome their lexical knowledge regarding the traits can apply to animate and inanimate entities.

\begin{figure}
    \fbox{\footnotesize\parbox{.465\textwidth}{
    A lucky fellow/peanut had a big \textbf{smile [1]} on his face. The fellow/peanut was \textbf{amazed [2]} about his good fortune. Just now he had won the jackpot of two million dollars. The fellow/peanut was elated/salted and who could blame him.
    }}
    \caption{Story from \citet{boudewyn2019adaptation}. N400 responses were recorded at the words in bold.}
    \label{fig:bdw_exp}
\end{figure}

\subsection{Adaptation Experiment}\label{sec:bdw} 
We now replicate \citeauthor{boudewyn2019adaptation}'s adaptation experiment, which combines the strengths of both prior experiments. Like the first, it captures adaptation over time; like the second, it avoids the potential issues of repetition.

\paragraph{Original Study} \citeauthor{boudewyn2019adaptation}'s adaptation experiment parallels the two previous experiments. Participants listened to 120 English-language stories containing either a typically or atypically animate entity (\Cref{fig:bdw_exp}). Participants' N400 responses were measured at the first content verb of the first two sentences of each story. These verbs signal that their subject is (perhaps atypically) animate, although they are notably not the same in each sentence. 
Findings mirrored those of \citeauthor{Nieuwland2006WhenPF}: there was a sharp drop in N400 response at between the two timesteps in the inanimate scenario. 
\paragraph{Our Experiment} For each of the 120 stories, we calculate the surprisal at the two critical verbs, in the animate and inanimate case.

\begin{figure}
    \includegraphics[width=\linewidth]{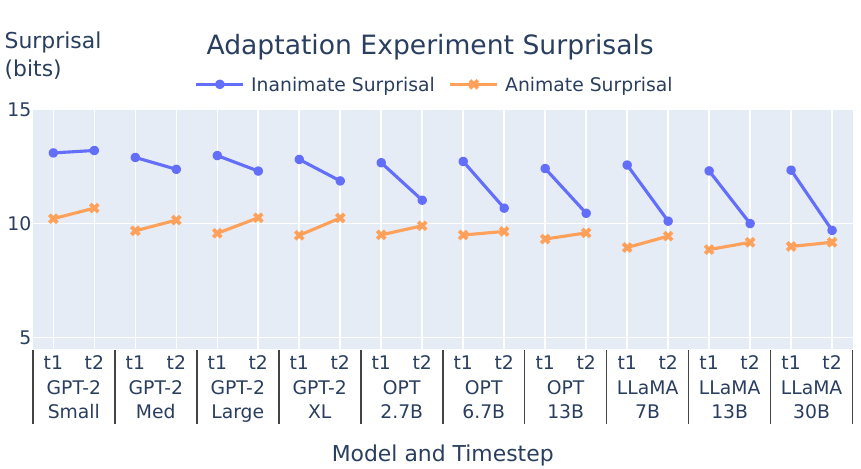}
    \caption{Adaptation experiment surprisals. Inanimate surprisal starts higher; the gap shrinks for larger LMs.}
    \label{fig:bdw_exp1_bar}
    \vspace{-5pt}
\end{figure}
\paragraph{Results}
The results of the adaptation experiment (\Cref{fig:bdw_exp1_bar}) might appear starkly different from those of the repetition experiment. As before, surprisal at the critical word drops in the inanimate case, though only slightly. But in the animate case, surprisal remains constant or increases. 

These results are in fact consistent with our earlier findings. The inanimate surprisal drop indicates that the short context sufficed to convince models of the entity's animacy. Moreover, the fact that surprisal does not drop in the animate case suggests that models are reacting specifically to the contextual cues that the inanimate entity is animate, as opposed to the context more generally. The importance of model size is also consistent: stronger models have a smaller gap between animate and inanimate surprisals at T2. Although this gap is still significant for current models, trends indicate that stronger models may eliminate it.

\subsection{Discussion}
Across experiments, models replicate broad trends in human N400 responses. 
Stronger models replicate human results better, with lower surprisals at inanimate entities. Do they thus process animacy more like humans? We caution that low surprisals are not always desirable for cognitive modeling, as surprisal from LMs can underestimate processing difficulty in terms of reading time \citep{vanschijndel2021single,arehalli-etal-2022-syntactic,oh2023surprisal}; this may be because even relatively small LMs can predict next words as well as humans \citep{goldstein2022shared}. Stronger LMs may only be better models of animacy processing in this situation because lower surprisals are desirable.

Regardless of the effects of model size, another question remains: do these positive results indicate that these LMs understand animacy? We cannot be certain: there exist mechanisms by which transformer LMs could perform well without any deep understanding of animacy. In the repetition experiment, models could use a copying mechanism to reduce their surprisal at repeated entities. In the context and even adaptation experiment, models could rely on the context, while ignoring the inanimate entity. This is a real concern: \citet{michaelov2023peanuts} construct a (simple) model that does this. 

In both cases, context is the complicating factor: LMs might exploit shallow context cues to simulate animacy processing effects, without having any real internal model of animacy. To investigate this question further, we study LMs' reactions to atypically animate entities in a low-context setting.

\section{Low-Context Atypical Animacy}\label{sec:low-context}
\begin{table*}
    \centering
    \begin{tabular}{c|c|c|c|c|c|c}
        Sentence & Rank & \#1& \#2& \#3& \#4& \#5\\
        \hline
         The kilt commented and started to &1&walk & get & move & laugh & say\\
         The cart noticed and was very &3&angry & ups& excited & interested & nerv\\
         The dime retired and was very&9994& rare & valuable & scar & popular & collect\\
         The telephone sighed and began to &10000& ring & v& bu& speak & be
     \end{tabular}
    \caption{Dataset examples and their top-5 continuations (sometimes partial words). The example at rank $n$ has the $n$th lowest animacy divergence (of 10,000). Low-divergence examples have animate continuations; high-divergence ones are stereotypical and inanimate. The top two examples use psychological verbs; the bottom two, physical.}
    \label{tab:exp3_examples}
\end{table*}
The previous experiments have shown that LMs can adapt in scenarios with atypically animate entities; however, LMs could have exploited shallow context features to do so, without any specific understanding or representation of animacy. We now investigate the extent to which LMs can leverage cues in the context by testing their behavior on very short sentences exhibiting atypical animacy.

\paragraph{Dataset}
We craft a set of short incomplete sentences that describe an atypically animate entity (\Cref{tab:exp3_examples}). The sentences are designed to elicit a critical next word---an adjective or a verb---that indicates if the LM treats the entity as animate. For example, if an LM continues ``The boat snored and started to'' with the verb ``dream'' this indicates that the boat is animate; continuing with ``sink'' does not. Unlike in prior experiments, these sentences provide only one clue indicating atypical animacy.

We create this dataset by defining prompts and filling them with nouns and verbs we sample from a predefined set. We sample from 181 nouns that humans rated as not very animate, but highly concrete; non-concrete inanimate nouns (e.g. ``fear'') cannot become animate, except metaphorically. We use concreteness ratings from \citet{wilson1988mrc}, and animacy ratings from \citet{vanarsdall2022animacy}. 

For the verbs, we use a manually-filtered set of 191 verbs that imply their subject is animate, from \citet{ji2018animacy}. Each verb implies that its subject is animate for either psychological or physical reasons; e.g. \textit{think} is psychological while \textit{walk} is physical. Each verb co-occurs with human subjects at a high, high-mid, or mid frequency. We create a dataset of 10,000 items (\Cref{tab:exp3_examples}) by sampling prompts, nouns, and verbs.\footnote{A list of prompts, nouns and verbs is in \Cref{app:low-context-dataset}. Full dataset is available at \url{https://github.com/hannamw/lms-in-love}. For further experiments that vary properties of this dataset, see \Cref{app:low-context-experiments}.}

\paragraph{Experiment} We run all LMs on the dataset. For each example, we evaluate whether the LM treats the entity in that example as animate by comparing the LM's output distribution to reference distributions. If our original sentence is $O=$``The chair spoke and began to'', our inanimate reference is $I=$``The chair began to'', while our animate reference is $A=$``The \texttt{[human]} began to'', with a human entity randomly sampled from \textit{person}, \textit{man}, \textit{woman}, \textit{boy}, \textit{girl}, and \textit{child}. We indicate via $D_{KL}(A||O)$ the divergence between the next-word distributions given $A$ as context, and given $O$ as context, with other KL divergences defined analogously. We focus in particular on $D_{KL}(A||O)$ as \textbf{animacy divergence}; lower animacy divergence implies a more ``animate'' distribution.

\begin{figure}
    \centering
    \includegraphics[width=\linewidth]{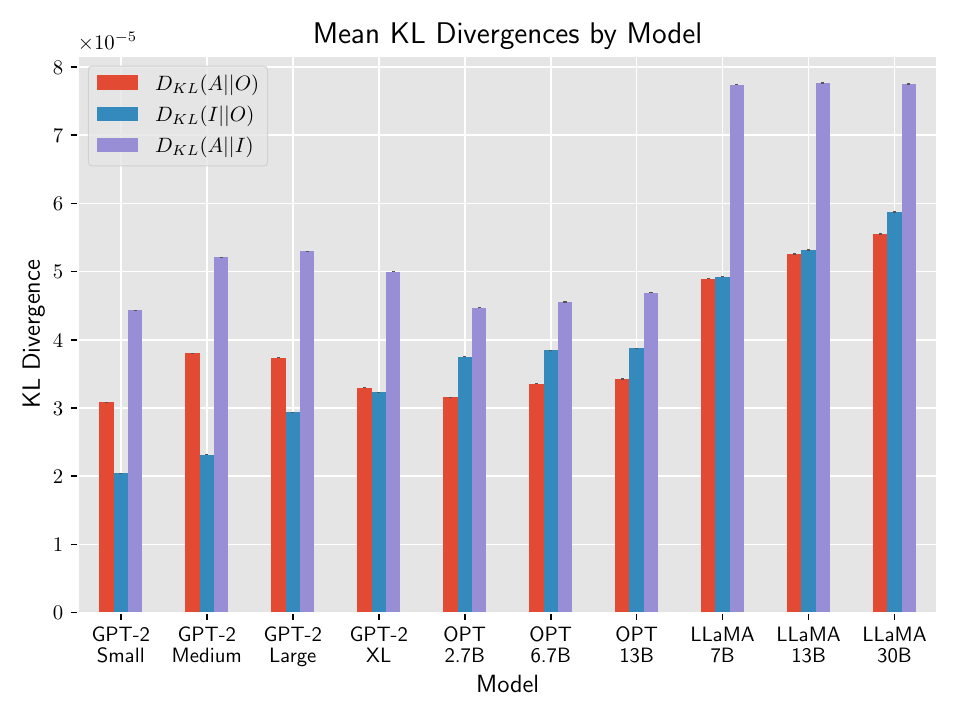}
    \caption{KL Divergence between atypically animate ($O$) and animate ($A$) / inanimate ($I$) references. Error bars (95\% CI) are marked, but extremely small. The lower the bar, the more similar the distributions. }
    \vspace{-5pt}
    \label{fig:exp3-bars}
\end{figure}

\begin{figure*}
    \centering
    \includegraphics[width=0.6\linewidth]{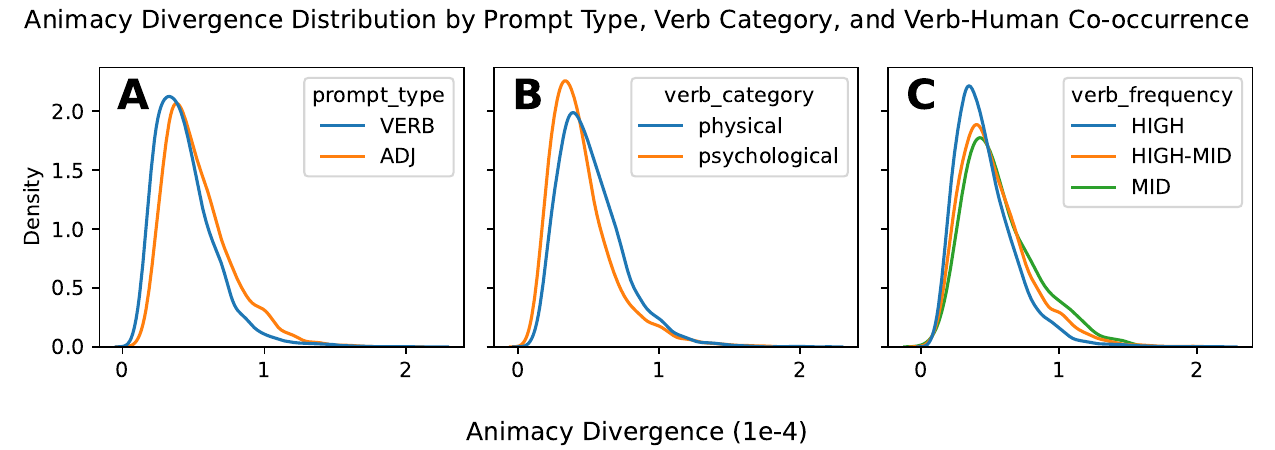}
    \includegraphics[width=0.39\linewidth]{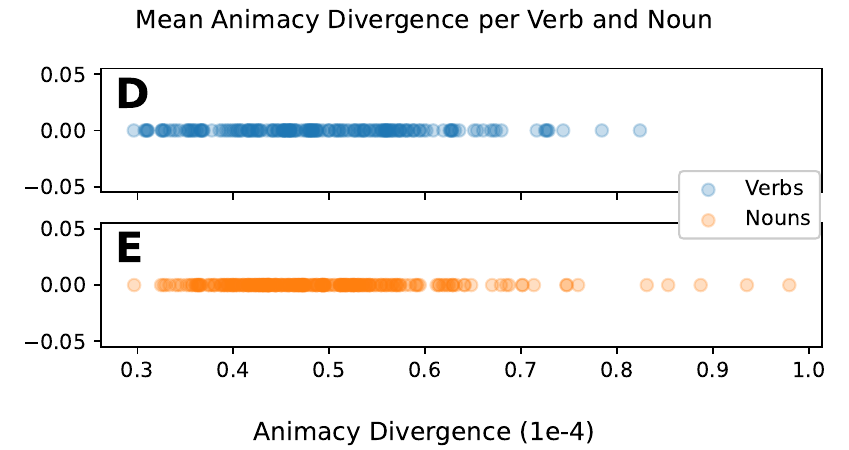}
    \caption{Left: Distribution of animacy divergences by prompt type, verb category, and verb-human co-occurrence. Right: Distribution of mean animacy divergences per-verb and per-noun. Each point is one verb (or noun).}
    \label{fig:distributions}
\end{figure*}

\paragraph{Quantitative Analysis} \Cref{fig:exp3-bars} shows KL divergences between the atypically animate sentence ($O$) and the reference distributions ($A/I$). For all models, the divergence between the inanimate and animate references (purple) is the highest. Although only the added verb separates the atypically animate sentence from the inanimate reference, this leads its divergence with the animate reference (red) to be consistently lower; that is, adding the verb significantly increased the distribution's animacy. For OPT models, the animacy divergence is the lowest of their three divergences: the atypically animate distribution is even more similar to the animate distribution (red) than inanimate distribution (blue). This trend holds true for LLaMA models as well, though only weakly. Still, all model behavior clearly shifts with the addition of the animacy-implying verb.

To understand the cause of this shift, we analyze the data with respect to the known factors that vary between our prompts, to discern which affected model behavior. Results were similar across models, so we display results for one model, \texttt{LLaMA-7B}. 

We first analyze the effect of our prompts, focusing on the difference between those that elicit verbs (``and began to\ldots'') and those that elicit adjectives (``and was very\ldots''). We find (\Cref{fig:distributions}, A) that verb-eliciting prompts produce lower animacy divergences than those that elicit adjectives.

We then analyze the effects of verbs and nouns on the sentence. For each verb or noun, we calculate the mean animacy divergence of sentences containing it. We observe a wide spread per verb and per noun (\Cref{fig:distributions}, D and E), suggesting that they both impact model behavior. We then study the factors affecting individual nouns' and verbs' divergences. We find that psychologically animate verbs have somewhat lower divergence than physically animate verbs; psychologically animate verbs produce more animate behavior from LMs ($p<0.01$, T-test; \Cref{fig:distributions} B). Sentences with verbs that had higher co-occurrence with humans had a lower divergence than those with a lower co-occurrence ($p<0.01$; \Cref{fig:distributions} C),\footnote{Significant for all three groups (F-test), and pairwise.} though the effect size is rather small. For the nouns, however, neither animacy nor concreteness explains trends in animacy divergence.

\paragraph{Qualitative Analysis} We also qualitatively verify that sentences with low animacy divergence have more animate continuations than those with high divergence. We sort examples by divergence and examine their top-5 continuations, focusing on examples at the top and bottom of the list.

\Cref{tab:exp3_examples} shows that animacy divergence aligns well with the qualitative animacy of continuations. Low-divergence examples have entities that \textit{laugh}, or are \textit{angry}. High-divergence ones have continuations stereotypical for the inanimate entity: a collectable \textit{dime} becomes \textit{rare}, or a \textit{telephone} begins to \textit{ring}.

\paragraph{Discussion} Results show that LMs can adapt to atypical animacy even given limited cues: one verb, rather than an entire story. They also suggest that LMs do not require a long context to adapt; however, the factors that regulate this adaptation are complex. Some, like the choice of prompt, bear no clear relation to animacy; others, like the nouns, show no clear pattern in how they affect model responses. Still, some interpretable factors exist. Psychological verbs may induce more animate continuations because they indicate animacy more strongly. While an inanimate object might metaphorically engage in physical activities like \textit{dance}, they seldom \textit{marry} or \textit{volunteer}. These psychological words may thus serve as stronger signals of animacy.

\section{Conclusions}
Although animacy manifests only indirectly in English---it is not morphologically marked---pre-trained LMs demonstrate relatively good animacy processing abilities. They both respect typical animacy and adapt to atypical animacy at close-to-human levels, although differences remain. They also demonstrate some ability to adapt to atypical animacy even when indicated by a very short context. LM adaptation still lags behind that of humans, but large models increasingly shrink the gap. We conclude that in the scenarios we test, LMs respond to animacy like humans do; however, our behavioral methodology can yield no conclusions about how models achieve this. Related work on world models \citep{li-etal-2021-implicit,li2023emergent} suggests that by using causal techniques to search for internal structure in models, future work could not only demonstrate that LMs respond well to animacy, but also explain how they do so.

\section*{Limitations}
In this study, we use primarily behavioral experiments. These are suitable for comparing models to human data, but do not reveal the causal mechanisms by which LMs process animacy. In order to discover these, it would be more appropriate to use causal interventions or similar techniques, which we do not explore.

Considering the limitations of behavioral techniques, this study is still limited by the fact that it did not collect human data. We translate \citeauthor{Nieuwland2006WhenPF}'s stimuli to English, but do not test them again on native English speakers; the N400 responses to Dutch data could differ from N400 responses to English data, even though we assume they will be similar. Similarly, studying human responses to our low-context atypical animacy stimuli (\Cref{sec:low-context}) would better inform our analysis of LM performance.

\section*{Ethics Statement}
This work presents only minor ethical concerns. A particular concern is one of bias and stereotypes; the original stories in \citet{Nieuwland2006WhenPF} do contain stereotypes. We attempt to soften these in our translations, but some stereotypes are still present in the translated material.

More generally, LMs such as those analyzed must be used with caution. Although such models achieve high performance on language-based tasks, this performance does not necessarily stem from genuine linguistic understanding. Moreover, models can not only perpetuate harmful biases present in their training data, but also create misleading or false output. 

\section*{Acknowledgements}
The authors thank the members of the Dialogue Modelling Group, particularly Joris Baan and Raquel Fern\'{a}ndez, as well as Marianne de Heer Kloots, for their helpful feedback. They also thank the authors of the original human studies, in particular Mante Nieuwland  and Megan Boudewyn, who allowed us to translate / use their stimuli. Finally, the authors thank participants of the ELLIS, AI4media, and AIDA Symposium on Large Language and Foundation Models in Amsterdam for their insightful comments. This work was supported by the ISRAEL SCIENCE FOUNDATION (grant No. 448/20), Open Philanthropy, and an Azrieli Foundation Early Career Faculty Fellowship.

\bibliography{anthology,custom}

\begin{thebibliography}{57}
\expandafter\ifx\csname natexlab\endcsname\relax\def\natexlab#1{#1}\fi

\bibitem[{Arehalli et~al.(2022)Arehalli, Dillon, and Linzen}]{arehalli-etal-2022-syntactic}
Suhas Arehalli, Brian Dillon, and Tal Linzen. 2022.
\newblock \href {https://aclanthology.org/2022.conll-1.20} {Syntactic surprisal from neural models predicts, but underestimates, human processing difficulty from syntactic ambiguities}.
\newblock In \emph{Proceedings of the 26th Conference on Computational Natural Language Learning (CoNLL)}, pages 301--313, Abu Dhabi, United Arab Emirates (Hybrid). Association for Computational Linguistics.

\bibitem[{Aurnhammer and Frank(2018)}]{aurnhammer2018comparing}
Christoph Aurnhammer and Stefan~L Frank. 2018.
\newblock \href {https://doi.org/10.31234/osf.io/wec74} {Comparing gated and simple recurrent neural network architectures as models of human sentence processing}.

\bibitem[{Baker et~al.(1998)Baker, Fillmore, and Lowe}]{baker-etal-1998-berkeley-framenet}
Collin~F. Baker, Charles~J. Fillmore, and John~B. Lowe. 1998.
\newblock \href {https://doi.org/10.3115/980845.980860} {The {B}erkeley {F}rame{N}et project}.
\newblock In \emph{36th Annual Meeting of the Association for Computational Linguistics and 17th International Conference on Computational Linguistics, Volume 1}, pages 86--90, Montreal, Quebec, Canada. Association for Computational Linguistics.

\bibitem[{Boudewyn et~al.(2019)Boudewyn, Blalock, Long, and Swaab}]{boudewyn2019adaptation}
Megan Boudewyn, Adam Blalock, Debra Long, and Tamara Swaab. 2019.
\newblock \href {https://doi.org/10.3758/s13415-019-00735-x} {Adaptation to animacy violations during listening comprehension}.
\newblock \emph{Cognitive, Affective, \& Behavioral Neuroscience}, 19.

\bibitem[{Bowman and Chopra(2012)}]{bowman-chopra-2012-automatic}
Samuel~R. Bowman and Harshit Chopra. 2012.
\newblock \href {https://aclanthology.org/N12-2002} {Automatic {A}nimacy classification}.
\newblock In \emph{Proceedings of the {NAACL} {HLT} 2012 Student Research Workshop}, pages 7--10, Montr{\'e}al, Canada. Association for Computational Linguistics.

\bibitem[{Bresnan and Hay(2008)}]{bresnan2008animacy}
Joan Bresnan and Jennifer Hay. 2008.
\newblock \href {https://doi.org/https://doi.org/10.1016/j.lingua.2007.02.007} {Gradient grammar: An effect of animacy on the syntax of give in new zealand and american english}.
\newblock \emph{Lingua}, 118(2):245--259.
\newblock Animacy, Argument Structure, and Argument Encoding.

\bibitem[{Bugaiska et~al.(2019)Bugaiska, Grégoire, Camblats, Gelin, Méot, and Bonin}]{bugaiska2019animacy}
Aurélia Bugaiska, Laurent Grégoire, Anna-Malika Camblats, Margaux Gelin, Alain Méot, and Patrick Bonin. 2019.
\newblock \href {https://doi.org/10.1177/1747021818771514} {Animacy and attentional processes: Evidence from the stroop task}.
\newblock \emph{Quarterly Journal of Experimental Psychology}, 72(4):882--889.
\newblock PMID: 29716460.

\bibitem[{Caplan et~al.(1994)Caplan, Hildebrandt, and Waters}]{caplan1994selectional}
David Caplan, Nancy Hildebrandt, and Gloria~S. Waters. 1994.
\newblock \href {https://doi.org/10.1080/01690969408402131} {Interaction of verb selectional restrictions, noun animacy and syntactic form in sentence processing}.
\newblock \emph{Language and Cognitive Processes}, 9(4):549--585.

\bibitem[{Caramazza and Shelton(1998)}]{caramazza1998domain}
Alfonso Caramazza and Jennifer~R. Shelton. 1998.
\newblock \href {https://doi.org/10.1162/089892998563752} {Domain-specific knowledge systems in the brain: The animate-inanimate distinction}.
\newblock \emph{J. Cognitive Neuroscience}, 10(1):1–34.

\bibitem[{Coll~Ardanuy et~al.(2020)Coll~Ardanuy, Nanni, Beelen, Hosseini, Ahnert, Lawrence, McDonough, Tolfo, Wilson, and McGillivray}]{coll-ardanuy-etal-2020-living}
Mariona Coll~Ardanuy, Federico Nanni, Kaspar Beelen, Kasra Hosseini, Ruth Ahnert, Jon Lawrence, Katherine McDonough, Giorgia Tolfo, Daniel~CS Wilson, and Barbara McGillivray. 2020.
\newblock \href {https://doi.org/10.18653/v1/2020.coling-main.400} {Living machines: A study of atypical animacy}.
\newblock In \emph{Proceedings of the 28th International Conference on Computational Linguistics}, pages 4534--4545, Barcelona, Spain (Online). International Committee on Computational Linguistics.

\bibitem[{Comrie(1989)}]{comrie1989language}
Bernard Comrie. 1989.
\newblock \emph{Language universals and linguistic typology: Syntax and morphology}.
\newblock University of Chicago press.

\bibitem[{de~Swart and de~Hoop(2018)}]{deSwart2018animacy}
Peter de~Swart and Helen de~Hoop. 2018.
\newblock \href {https://doi.org/doi:10.1515/tl-2018-0001} {Shifting animacy}.
\newblock \emph{Theoretical Linguistics}, 44(1-2):1--23.

\bibitem[{de~Vries and Nissim(2021)}]{de-vries-nissim-2021-good}
Wietse de~Vries and Malvina Nissim. 2021.
\newblock \href {https://doi.org/10.18653/v1/2021.findings-acl.74} {As good as new. how to successfully recycle {E}nglish {GPT}-2 to make models for other languages}.
\newblock In \emph{Findings of the Association for Computational Linguistics: ACL-IJCNLP 2021}, pages 836--846, Online. Association for Computational Linguistics.

\bibitem[{Elman(1990)}]{elman1990finding}
Jeffrey~L. Elman. 1990.
\newblock \href {https://doi.org/https://doi.org/10.1207/s15516709cog1402\_1} {Finding structure in time}.
\newblock \emph{Cognitive Science}, 14(2):179--211.

\bibitem[{Ettinger(2020)}]{ettinger-2020-bert}
Allyson Ettinger. 2020.
\newblock \href {https://doi.org/10.1162/tacl_a_00298} {What {BERT} is not: Lessons from a new suite of psycholinguistic diagnostics for language models}.
\newblock \emph{Transactions of the Association for Computational Linguistics}, 8:34--48.

\bibitem[{Ferreira(1994)}]{ferreira1994passive}
F.~Ferreira. 1994.
\newblock \href {https://doi.org/https://doi.org/10.1006/jmla.1994.1034} {Choice of passive voice is affected by verb type and animacy}.
\newblock \emph{Journal of Memory and Language}, 33(6):715--736.

\bibitem[{Frank et~al.(2013)Frank, Otten, Galli, and Vigliocco}]{frank-etal-2013-word}
Stefan~L. Frank, Leun~J. Otten, Giulia Galli, and Gabriella Vigliocco. 2013.
\newblock \href {https://aclanthology.org/P13-2152} {Word surprisal predicts n400 amplitude during reading}.
\newblock In \emph{Proceedings of the 51st Annual Meeting of the Association for Computational Linguistics (Volume 2: Short Papers)}, pages 878--883, Sofia, Bulgaria. Association for Computational Linguistics.

\bibitem[{Frank et~al.(2015)Frank, Otten, Galli, and Vigliocco}]{FRANK20151}
Stefan~L. Frank, Leun~J. Otten, Giulia Galli, and Gabriella Vigliocco. 2015.
\newblock \href {https://doi.org/https://doi.org/10.1016/j.bandl.2014.10.006} {The {ERP} response to the amount of information conveyed by words in sentences}.
\newblock \emph{Brain and Language}, 140:1--11.

\bibitem[{Goldhahn et~al.(2012)Goldhahn, Eckart, and Quasthoff}]{goldhahn-etal-2012-building}
Dirk Goldhahn, Thomas Eckart, and Uwe Quasthoff. 2012.
\newblock \href {http://www.lrec-conf.org/proceedings/lrec2012/pdf/327_Paper.pdf} {Building large monolingual dictionaries at the {L}eipzig corpora collection: From 100 to 200 languages}.
\newblock In \emph{Proceedings of the Eighth International Conference on Language Resources and Evaluation ({LREC}'12)}, pages 759--765, Istanbul, Turkey. European Language Resources Association (ELRA).

\bibitem[{Goldstein et~al.(2022)Goldstein, Zada, Buchnik, Schain, Price, Aubrey, Nastase, Feder, Emanuel, Cohen, Jansen, Gazula, Choe, Rao, Kim, Casto, Fanda, Doyle, Friedman, Dugan, Melloni, Reichart, Devore, Flinker, Hasenfratz, Levy, Hassidim, Brenner, Matias, Norman, Devinsky, and Uri}]{goldstein2022shared}
Ariel Goldstein, Zaid Zada, Eliav Buchnik, Mariano Schain, Amy Price, Bobbi Aubrey, Samuel~A. Nastase, Amir Feder, Dotan Emanuel, Alon Cohen, Aren Jansen, Harshvardhan Gazula, Gina Choe, Aditi Rao, Catherine Kim, Colton Casto, Lora Fanda, Werner Doyle, Daniel Friedman, Patricia Dugan, Lucia Melloni, Roi Reichart, Sasha Devore, Adeen Flinker, Liat Hasenfratz, Omer Levy, Avinatan Hassidim, Michael Brenner, Yossi Matias, Kenneth~A. Norman, Orrin Devinsky, and Hasson Uri. 2022.
\newblock \href {https://doi.org/10.1038/s41593-022-01026-4} {Shared computational principles for language processing in humans and deep language models}.
\newblock \emph{Nature Neuroscience}, 25:369--380.

\bibitem[{Goodkind and Bicknell(2018)}]{goodkind-bicknell-2018-predictive}
Adam Goodkind and Klinton Bicknell. 2018.
\newblock \href {https://doi.org/10.18653/v1/W18-0102} {Predictive power of word surprisal for reading times is a linear function of language model quality}.
\newblock In \emph{Proceedings of the 8th Workshop on Cognitive Modeling and Computational Linguistics ({CMCL} 2018)}, pages 10--18, Salt Lake City, Utah. Association for Computational Linguistics.

\bibitem[{Havinga(2021)}]{yhavingaDutchGPT2}
Yeb Havinga. 2021.
\newblock \href {https://huggingface.co/yhavinga/gpt2-medium-dutch} {{GPT}-2-medium-{D}utch}.

\bibitem[{Jahan et~al.(2018)Jahan, Chauhan, and Finlayson}]{jahan-etal-2018-new}
Labiba Jahan, Geeticka Chauhan, and Mark Finlayson. 2018.
\newblock \href {https://aclanthology.org/C18-1001} {A new approach to {A}nimacy detection}.
\newblock In \emph{Proceedings of the 27th International Conference on Computational Linguistics}, pages 1--12, Santa Fe, New Mexico, USA. Association for Computational Linguistics.

\bibitem[{Ji and Liang(2018)}]{ji2018animacy}
Jie Ji and Maocheng Liang. 2018.
\newblock \href {https://doi.org/https://doi.org/10.1016/j.lingua.2017.12.017} {An animacy hierarchy within inanimate nouns: English corpus evidence from a prototypical perspective}.
\newblock \emph{Lingua}, 205:71--89.

\bibitem[{Karsdorp et~al.(2015)Karsdorp, van~der Meulen, Meder, and van~den Bosch}]{karsdorp2015animacy}
Folgert Karsdorp, Marten van~der Meulen, Theo Meder, and Antal van~den Bosch. 2015.
\newblock \href {https://doi.org/10.4230/OASIcs.CMN.2015.82} {{Animacy Detection in Stories}}.
\newblock In \emph{6th Workshop on Computational Models of Narrative (CMN 2015)}, volume~45 of \emph{OpenAccess Series in Informatics (OASIcs)}, pages 82--97, Dagstuhl, Germany. Schloss Dagstuhl--Leibniz-Zentrum fuer Informatik.

\bibitem[{Kauf et~al.(2022)Kauf, Ivanova, Rambelli, Chersoni, She, Chowdhury, Fedorenko, and Lenci}]{kauf2022event}
Carina Kauf, Anna~A. Ivanova, Giulia Rambelli, Emmanuele Chersoni, Jingyuan~S. She, Zawad Chowdhury, Evelina Fedorenko, and Alessandro Lenci. 2022.
\newblock \href {http://arxiv.org/abs/2212.01488} {Event knowledge in large language models: the gap between the impossible and the unlikely}.

\bibitem[{Kipper et~al.(2000)Kipper, Dang, and Palmer}]{kipper2000verbnet}
Karin Kipper, Hoa~Trang Dang, and Martha Palmer. 2000.
\newblock Class-based construction of a verb lexicon.
\newblock In \emph{Proceedings of the Seventeenth National Conference on Artificial Intelligence and Twelfth Conference on Innovative Applications of Artificial Intelligence}, page 691–696. AAAI Press.

\bibitem[{Kuno and Kaburaki(1977)}]{kuno1977empathy}
Susumu Kuno and Etsuko Kaburaki. 1977.
\newblock Empathy and syntax.
\newblock \emph{Linguistic inquiry}, pages 627--672.

\bibitem[{Li et~al.(2021)Li, Nye, and Andreas}]{li-etal-2021-implicit}
Belinda~Z. Li, Maxwell Nye, and Jacob Andreas. 2021.
\newblock \href {https://doi.org/10.18653/v1/2021.acl-long.143} {Implicit representations of meaning in neural language models}.
\newblock In \emph{Proceedings of the 59th Annual Meeting of the Association for Computational Linguistics and the 11th International Joint Conference on Natural Language Processing (Volume 1: Long Papers)}, pages 1813--1827, Online. Association for Computational Linguistics.

\bibitem[{Li et~al.(2023)Li, Hopkins, Bau, Vi{\'e}gas, Pfister, and Wattenberg}]{li2023emergent}
Kenneth Li, Aspen~K Hopkins, David Bau, Fernanda Vi{\'e}gas, Hanspeter Pfister, and Martin Wattenberg. 2023.
\newblock \href {https://openreview.net/forum?id=DeG07_TcZvT} {Emergent world representations: Exploring a sequence model trained on a synthetic task}.
\newblock In \emph{The Eleventh International Conference on Learning Representations}.

\bibitem[{Linzen et~al.(2016)Linzen, Dupoux, and Goldberg}]{linzen-etal-2016-assessing}
Tal Linzen, Emmanuel Dupoux, and Yoav Goldberg. 2016.
\newblock \href {https://doi.org/10.1162/tacl_a_00115} {Assessing the ability of {LSTM}s to learn syntax-sensitive dependencies}.
\newblock \emph{Transactions of the Association for Computational Linguistics}, 4:521--535.

\bibitem[{Merkx and Frank(2021)}]{merkx-frank-2021-human}
Danny Merkx and Stefan~L. Frank. 2021.
\newblock \href {https://doi.org/10.18653/v1/2021.cmcl-1.2} {Human sentence processing: Recurrence or attention?}
\newblock In \emph{Proceedings of the Workshop on Cognitive Modeling and Computational Linguistics}, pages 12--22, Online. Association for Computational Linguistics.

\bibitem[{Michaelov and Bergen(2020)}]{michaelov-bergen-2020-well}
James Michaelov and Benjamin Bergen. 2020.
\newblock \href {https://doi.org/10.18653/v1/2020.conll-1.53} {How well does surprisal explain n400 amplitude under different experimental conditions?}
\newblock In \emph{Proceedings of the 24th Conference on Computational Natural Language Learning}, pages 652--663, Online. Association for Computational Linguistics.

\bibitem[{Michaelov et~al.(2021)Michaelov, Bardolph, Coulson, and Bergen}]{michaelov2021different}
James~A. Michaelov, Megan~D. Bardolph, Seana Coulson, and Benjamin~K. Bergen. 2021.
\newblock \href {https://escholarship.org/content/qt9z06m20f/qt9z06m20f_noSplash_531cb01b98cc5f2c87a64d1bd65b63c9.pdf?t=qwi3v7} {Different kinds of cognitive plausibility: why are transformers better than {RNNs} at predicting {N}400 amplitude?}
\newblock In \emph{Proceedings of the 43rd Annual Meeting of the Cognitive Science Society}, pages 300--306.

\bibitem[{Michaelov et~al.(2022)Michaelov, Coulson, and Bergen}]{michaelov2022n400}
James~A. Michaelov, Seana Coulson, and Benjamin~K. Bergen. 2022.
\newblock \href {https://doi.org/10.1109/TCDS.2022.3176783} {So cloze yet so far: N400 amplitude is better predicted by distributional information than human predictability judgements}.
\newblock \emph{IEEE Transactions on Cognitive and Developmental Systems}, pages 1--1.

\bibitem[{Michaelov et~al.(2023)Michaelov, Coulson, and Bergen}]{michaelov2023peanuts}
James~A. Michaelov, Seana Coulson, and Benjamin~K. Bergen. 2023.
\newblock \href {http://arxiv.org/abs/2301.08731} {Can peanuts fall in love with distributional semantics?}
\newblock To appear in the Proceedings of the 45th Annual Meeting of the Cognitive Science Society (Sydney, Australia; 2023).

\bibitem[{Nairne et~al.(2013)Nairne, VanArsdall, Pandeirada, Cogdill, and LeBreton}]{nairne2013adaptive}
James~S. Nairne, Joshua~E. VanArsdall, Josefa N.~S. Pandeirada, Mindi Cogdill, and James~M. LeBreton. 2013.
\newblock \href {http://www.jstor.org/stable/24539404} {Adaptive memory: The mnemonic value of animacy}.
\newblock \emph{Psychological Science}, 24(10):2099--2105.

\bibitem[{New et~al.(2007)New, Cosmides, and Tooby}]{new2007category}
Joshua New, Leda Cosmides, and John Tooby. 2007.
\newblock \href {https://doi.org/10.1073/pnas.0703913104} {Category-specific attention for animals reflects ancestral priorities, not expertise}.
\newblock \emph{Proceedings of the National Academy of Sciences}, 104(42):16598--16603.

\bibitem[{Nieuwland and van Berkum(2006)}]{Nieuwland2006WhenPF}
Mante~S. Nieuwland and Jos J.~A. van Berkum. 2006.
\newblock \href {https://doi.org/10.1162/jocn.2006.18.7.1098} {When peanuts fall in love: N400 evidence for the power of discourse}.
\newblock \emph{J. Cognitive Neuroscience}, 18(7):1098–1111.

\bibitem[{Oh and Schuler(2023)}]{oh2023surprisal}
Byung-Doh Oh and William Schuler. 2023.
\newblock \href {https://doi.org/10.1162/tacl_a_00548} {{Why Does Surprisal From Larger Transformer-Based Language Models Provide a Poorer Fit to Human Reading Times?}}
\newblock \emph{Transactions of the Association for Computational Linguistics}, 11:336--350.

\bibitem[{Olsson et~al.(2022)Olsson, Elhage, Nanda, Joseph, DasSarma, Henighan, Mann, Askell, Bai, Chen, Conerly, Drain, Ganguli, Hatfield-Dodds, Hernandez, Johnston, Jones, Kernion, Lovitt, Ndousse, Amodei, Brown, Clark, Kaplan, McCandlish, and Olah}]{olsson2022context}
Catherine Olsson, Nelson Elhage, Neel Nanda, Nicholas Joseph, Nova DasSarma, Tom Henighan, Ben Mann, Amanda Askell, Yuntao Bai, Anna Chen, Tom Conerly, Dawn Drain, Deep Ganguli, Zac Hatfield-Dodds, Danny Hernandez, Scott Johnston, Andy Jones, Jackson Kernion, Liane Lovitt, Kamal Ndousse, Dario Amodei, Tom Brown, Jack Clark, Jared Kaplan, Sam McCandlish, and Chris Olah. 2022.
\newblock \href {https://transformer-circuits.pub/2022/in-context-learning-and-induction-heads/index.html} {In-context learning and induction heads}.
\newblock \emph{Transformer Circuits Thread}.

\bibitem[{Orasan and Evans(2007)}]{orasan2007animacy}
Constantin Orasan and Richard Evans. 2007.
\newblock \href {https://dl.acm.org/doi/10.5555/1622606.1622610} {{NP} animacy identification for anaphora resolution}.
\newblock \emph{J. Artif. Int. Res.}, 29(1):79–103.

\bibitem[{Paszke et~al.(2019)Paszke, Gross, Massa, Lerer, Bradbury, Chanan, Killeen, Lin, Gimelshein, Antiga, Desmaison, Kopf, Yang, DeVito, Raison, Tejani, Chilamkurthy, Steiner, Fang, Bai, and Chintala}]{NEURIPS2019_9015}
Adam Paszke, Sam Gross, Francisco Massa, Adam Lerer, James Bradbury, Gregory Chanan, Trevor Killeen, Zeming Lin, Natalia Gimelshein, Luca Antiga, Alban Desmaison, Andreas Kopf, Edward Yang, Zachary DeVito, Martin Raison, Alykhan Tejani, Sasank Chilamkurthy, Benoit Steiner, Lu~Fang, Junjie Bai, and Soumith Chintala. 2019.
\newblock \href {http://papers.neurips.cc/paper/9015-pytorch-an-imperative-style-high-performance-deep-learning-library.pdf} {Pytorch: An imperative style, high-performance deep learning library}.
\newblock In H.~Wallach, H.~Larochelle, A.~Beygelzimer, F.~d\textquotesingle Alch\'{e}-Buc, E.~Fox, and R.~Garnett, editors, \emph{Advances in Neural Information Processing Systems 32}, pages 8024--8035. Curran Associates, Inc.

\bibitem[{Radford et~al.(2019)Radford, Wu, Child, Luan, Amodei, and Sutskever}]{radford2019language}
Alec Radford, Jeff Wu, Rewon Child, David Luan, Dario Amodei, and Ilya Sutskever. 2019.
\newblock \href {https://d4mucfpksywv.cloudfront.net/better-language-models/language_models_are_unsupervised_multitask_learners.pdf} {Language models are unsupervised multitask learners}.

\bibitem[{Rakison and Poulin-Dubois(2001)}]{rakison2021developmental}
David Rakison and Diane Poulin-Dubois. 2001.
\newblock \href {https://doi.org/10.1037/0033-2909.127.2.209} {Developmental origin of the animate-inanimate distinction}.
\newblock \emph{Psychological Bulletin}, 127:209--228.

\bibitem[{Rosenbach(2008)}]{rosenbach2008animacy}
Anette Rosenbach. 2008.
\newblock \href {https://doi.org/https://doi.org/10.1016/j.lingua.2007.02.002} {Animacy and grammatical variation—findings from english genitive variation}.
\newblock \emph{Lingua}, 118(2):151--171.
\newblock Animacy, Argument Structure, and Argument Encoding.

\bibitem[{Sinclair et~al.(2022)Sinclair, Jumelet, Zuidema, and Fern{\'a}ndez}]{sinclair-etal-2022-structural}
Arabella Sinclair, Jaap Jumelet, Willem Zuidema, and Raquel Fern{\'a}ndez. 2022.
\newblock \href {https://doi.org/10.1162/tacl_a_00504} {Structural persistence in language models: Priming as a window into abstract language representations}.
\newblock \emph{Transactions of the Association for Computational Linguistics}, 10:1031--1050.

\bibitem[{Smith and Levy(2013)}]{smith2013predictability}
Nathaniel~J. Smith and Roger Levy. 2013.
\newblock \href {https://doi.org/https://doi.org/10.1016/j.cognition.2013.02.013} {The effect of word predictability on reading time is logarithmic}.
\newblock \emph{Cognition}, 128(3):302--319.

\bibitem[{Touvron et~al.(2023)Touvron, Lavril, Izacard, Martinet, Lachaux, Lacroix, Rozi{\`e}re, Goyal, Hambro, Azhar, Rodriguez, Joulin, Grave, and Lample}]{touvron2023llama}
Hugo Touvron, Thibaut Lavril, Gautier Izacard, Xavier Martinet, Marie-Anne Lachaux, Timoth{\'e}e Lacroix, Baptiste Rozi{\`e}re, Naman Goyal, Eric Hambro, Faisal Azhar, Aurelien Rodriguez, Armand Joulin, Edouard Grave, and Guillaume Lample. 2023.
\newblock \href {https://arxiv.org/abs/2302.13971} {Llama: Open and efficient foundation language models}.
\newblock \emph{arXiv preprint arXiv:2302.13971}.

\bibitem[{van Schijndel and Linzen(2021)}]{vanschijndel2021single}
Marten van Schijndel and Tal Linzen. 2021.
\newblock \href {https://doi.org/https://doi.org/10.1111/cogs.12988} {Single-stage prediction models do not explain the magnitude of syntactic disambiguation difficulty}.
\newblock \emph{Cognitive Science}, 45(6):e12988.

\bibitem[{VanArsdall and Blunt(2022)}]{vanarsdall2022animacy}
Joshua~E. VanArsdall and Janell~R. Blunt. 2022.
\newblock \href {https://doi.org/10.3758/s13421-021-01266-y} {{Analyzing the structure of animacy: Exploring relationships among six new animacy and 15 existing normative dimensions for 1,200 concrete nouns}}.
\newblock \emph{Memory \& Cognition}, 50:997--1012.

\bibitem[{Warstadt et~al.(2020)Warstadt, Parrish, Liu, Mohananey, Peng, Wang, and Bowman}]{warstadt-etal-2020-blimp-benchmark}
Alex Warstadt, Alicia Parrish, Haokun Liu, Anhad Mohananey, Wei Peng, Sheng-Fu Wang, and Samuel~R. Bowman. 2020.
\newblock \href {https://doi.org/10.1162/tacl_a_00321} {{BL}i{MP}: The benchmark of linguistic minimal pairs for {E}nglish}.
\newblock \emph{Transactions of the Association for Computational Linguistics}, 8:377--392.

\bibitem[{Wilcox et~al.(2020)Wilcox, Gauthier, Hu, Qian, and Levy}]{wilcox2020predictive}
Ethan~Gotlieb Wilcox, Jon Gauthier, Jennifer Hu, Peng Qian, and Roger Levy. 2020.
\newblock \href {https://cognitivesciencesociety.org/cogsci20/papers/0375/0375.pdf} {On the predictive power of neural language models for human real-time comprehension behavior}.
\newblock In \emph{Proceedings of the 42nd Annual Meeting of the Cognitive Science Society}.

\bibitem[{Wilcoxon(1945)}]{wilcoxon1945ranking}
Frank Wilcoxon. 1945.
\newblock \href {http://www.jstor.org/stable/3001968} {Individual comparisons by ranking methods}.
\newblock \emph{Biometrics Bulletin}, 1(6):80--83.

\bibitem[{Wilson(1988)}]{wilson1988mrc}
Michael Wilson. 1988.
\newblock \href {https://doi.org/10.3758/BF03202594} {{MRC psycholinguistic database: Machine-usable dictionary, version 2.00}}.
\newblock \emph{Behavior Research Methods, Instruments, \& Computers}, 20:6--10.

\bibitem[{Wolf et~al.(2020)Wolf, Debut, Sanh, Chaumond, Delangue, Moi, Cistac, Rault, Louf, Funtowicz, Davison, Shleifer, von Platen, Ma, Jernite, Plu, Xu, Le~Scao, Gugger, Drame, Lhoest, and Rush}]{wolf-etal-2020-transformers}
Thomas Wolf, Lysandre Debut, Victor Sanh, Julien Chaumond, Clement Delangue, Anthony Moi, Pierric Cistac, Tim Rault, Remi Louf, Morgan Funtowicz, Joe Davison, Sam Shleifer, Patrick von Platen, Clara Ma, Yacine Jernite, Julien Plu, Canwen Xu, Teven Le~Scao, Sylvain Gugger, Mariama Drame, Quentin Lhoest, and Alexander Rush. 2020.
\newblock \href {https://doi.org/10.18653/v1/2020.emnlp-demos.6} {Transformers: State-of-the-art natural language processing}.
\newblock In \emph{Proceedings of the 2020 Conference on Empirical Methods in Natural Language Processing: System Demonstrations}, pages 38--45, Online. Association for Computational Linguistics.

\bibitem[{Zhang et~al.(2022)Zhang, Roller, Goyal, Artetxe, Chen, Chen, Dewan, Diab, Li, Lin, Mihaylov, Ott, Shleifer, Shuster, Simig, Koura, Sridhar, Wang, and Zettlemoyer}]{zhang2022OPT}
Susan Zhang, Stephen Roller, Naman Goyal, Mikel Artetxe, Moya Chen, Shuohui Chen, Christopher Dewan, Mona Diab, Xian Li, Xi~Victoria Lin, Todor Mihaylov, Myle Ott, Sam Shleifer, Kurt Shuster, Daniel Simig, Punit~Singh Koura, Anjali Sridhar, Tianlu Wang, and Luke Zettlemoyer. 2022.
\newblock \href {https://doi.org/10.48550/ARXIV.2205.01068} {{OPT}: Open pre-trained transformer language models}.

\end{thebibliography}
\bibliographystyle{acl_natbib}

\appendix

\section{Implementation Details}\label{app:implementation-details} 
We implement all experiments in PyTorch \citep{NEURIPS2019_9015}. For all models but LLaMA, we use the implementations and weights publicly available via the HuggingFace Transformers library \citep{wolf-etal-2020-transformers}; for LLaMA, weights are only available upon request via a form at \url{https://github.com/facebookresearch/llama}. We run models using an Nvidia A100 40GB GPU (or multiple when necessary, as for LLaMA 30B). The runtime of these  experiments should not exceed 24 hours, even when run serially.

\section{N400 Results for Dutch Data}\label{app:dutch-data}
The methods for this set of experiments are mostly identical to those of the English experiments. However, instead of English LMs, we use Dutch LMs. We consider GPT-2 Small trained from scratch on Dutch, and GPT-2 Medium trained in English, with fine-tuned Dutch word embeddings \citep{de-vries-nissim-2021-good}, and GPT-2 Medium and Large trained from scratch on Dutch \citep{yhavingaDutchGPT2}. These represent the best Dutch autoregressive LMs available.

\subsection{Results: Repetition Experiment}
\begin{figure}
    \includegraphics[width=\linewidth]{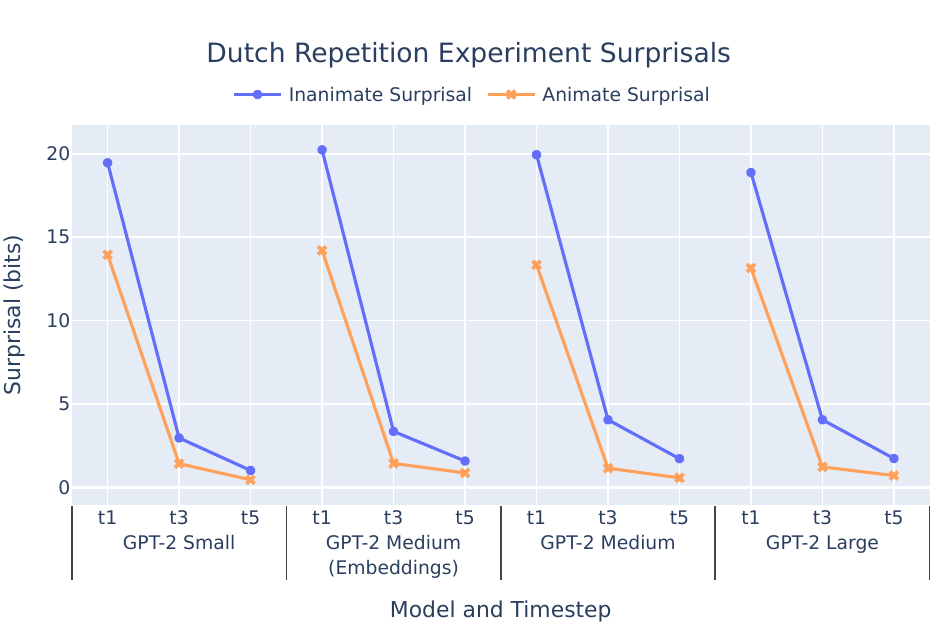}
    \caption{Repetition experiment surprisals. Surprisal drops rapidly after T1, with inanimate surprisal drawing close to animate surprisal.}
    \label{fig:nvb_exp1_line_dutch}
\end{figure}

The results of the repetition experiment in Dutch (\Cref{fig:nvb_exp1_line_dutch}) are rather similar to those in English. As before, the surprisal starts out high for both animate and inanimate entities (though higher for the latter). The difference between these two is less pronounced than in English. The surprisal drops rapidly at T3, and again at T5. Unlike in English, there are no strong model-wise trends, whereby stronger models have a lower difference in surprisals. And in all cases, the difference between the two conditions at T5 is statistically significant.

\begin{figure}
    \includegraphics[width=\linewidth]{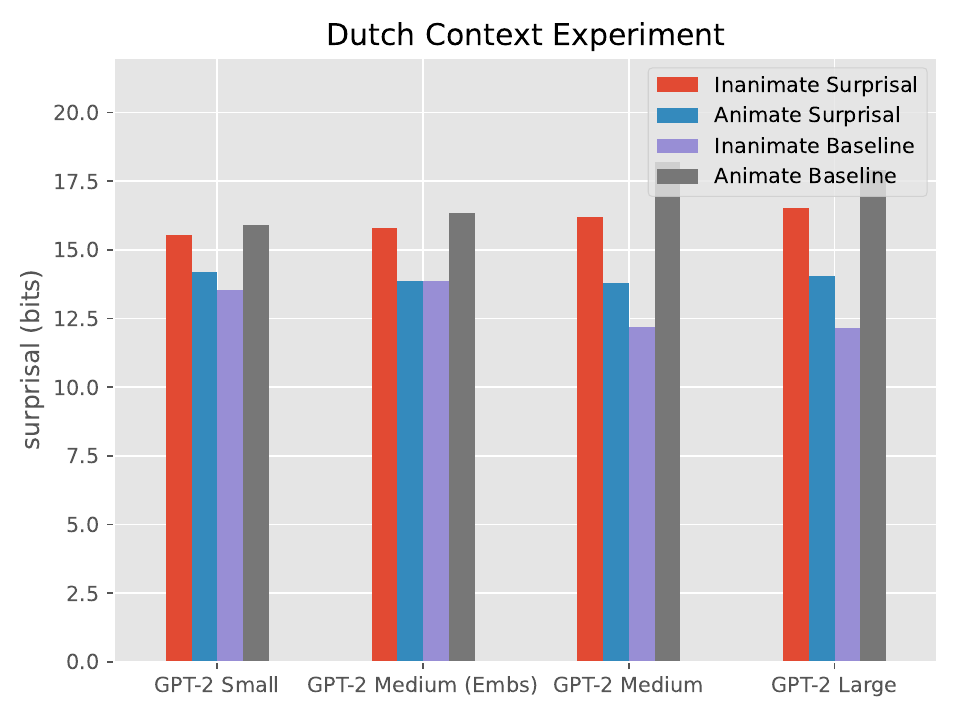}
    \caption{Context experiment surprisals. With context, the animate adjective is much less surprising; in the contextless baseline condition, this is reversed.}
    \label{fig:nvb_exp2_bar_dutch}
\end{figure}

\subsection{Results: Context Experiment}
Again, the results of the Dutch experiment (\Cref{fig:nvb_exp2_bar_dutch}) are much like the English results. Surprisals at animate adjectives are much lower than those at inanimate adjectives; however, in the baseline condition, which lacks context, the trend is reversed.

\section{\citeauthor{boudewyn2019adaptation}: English Context Experiment}\label{app:bdw_exp2}

\begin{figure}
    \fbox{\footnotesize\parbox{.465\textwidth}{
    A lucky peanut had a big smile on his face. The peanut was amazed about his good fortune. Just now he had won the jackpot of two million dollars. The peanut was \textbf{elated/salted} and who could blame him.
    }}
    \caption{Story from \citet{boudewyn2019adaptation}, context experiment. N400 responses were recorded at the words in bold.}
    \label{fig:bdw_exp2}
\end{figure}

\begin{figure}
    \includegraphics[width=\linewidth]{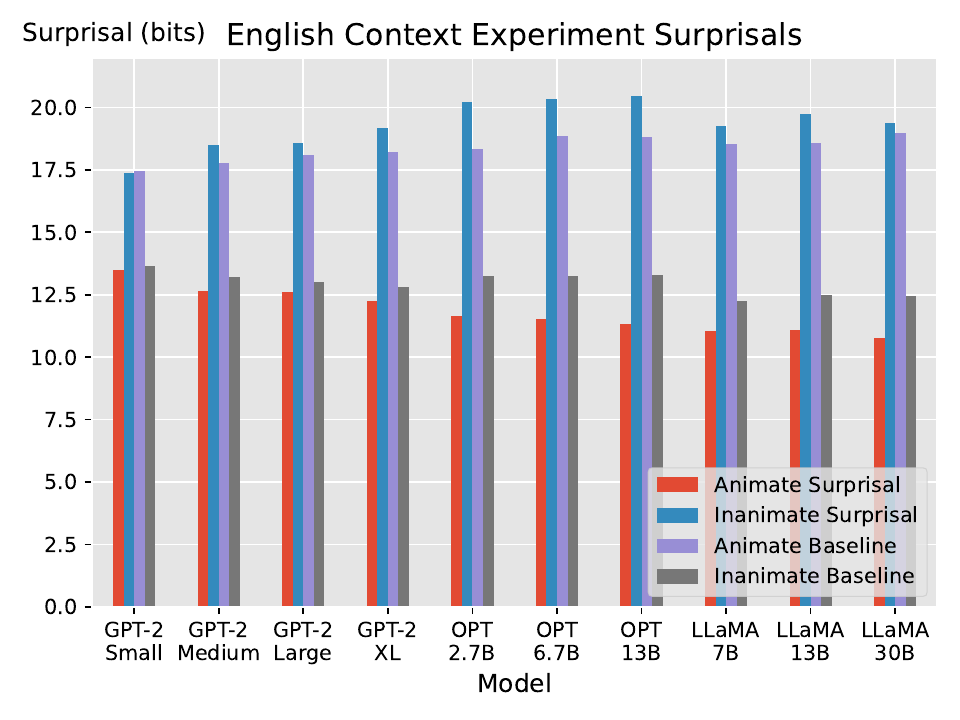}
    \caption{Surprisals for animate and inanimate adjectives in the normal and baseline condition}
    \label{fig:bdw_exp2_bar}
\end{figure}

\paragraph{Original Study}
Much like in \citet{Nieuwland2006WhenPF}, \citet{boudewyn2019adaptation} also measured participants' N400 responses to adjectives at the end of each story (\Cref{fig:bdw_exp2}). Each adjective was either typical for animate entities (and thus context-appropriate) or typical for the inanimate entity in question (but context-inappropriate). N400 responses were much lower in the former case than in the latter.

\paragraph{Experiment} As in the earlier context experiment (\Cref{sec:nvb_exp2}), for each of the 120 stories, we calculate the surprisal at the animate and inanimate adjective. We also compute baseline surprisals, defined as the surprisal of the critical adjective given only the sentence containing it as context.
\paragraph{Results}
The results of the English context experiment  (\Cref{fig:bdw_exp2_bar}) are like those of the earlier context experiment (\Cref{sec:nvb_exp2}). Like before, surprisal at the animate adjective is much lower than surprisal at the inanimate adjective; in the baseline condition, this trend is reversed.

\section{Further Low-Context Experiments}\label{app:low-context-experiments}
In the following experiments, we made changes to our and experimental setup, in order to ensure that our findings were not a result of any idiosyncrasies of our dataset's construction.

\subsection{Larger Human Entity Sample Pool Experiment}
In our original experiment, our human reference was $A=$``The \texttt{[human]} began to'', where \texttt{[human]} was sampled from a very limited pool: \textit{person}, \textit{man}, \textit{woman}, \textit{boy}, \textit{girl}, and \textit{child}. The pool was limited for two reasons. First, we did not want to introduce another axis of variation across examples. 

Second, and more importantly, we wanted to create a generic ``human-like'' action distribution. Some nouns (e.g. a musician, or a thief) are animate humans, but their distributions over next actions are probably skewed in ways that are not representative of animate entities in general; musicians are likely to play music, and thieves, to steal. Thus, their next-token distributions might have a high KL-divergence with those of atypically animate objects for reasons unrelated to the object's perceived animacy. However, we still wanted to ensure that our findings are not reliant on this small pool of humans. To verify this, we constructed a larger pool of human entities to sample from. 

To construct this pool, we first added in generic human entities (\textit{man}, \textit{woman}, \textit{person}, \textit{boy}, \textit{girl}, \textit{child}, \textit{teenager}) and family relations (\textit{mother}, \textit{father}, \textit{grandfather}, \textit{grandmother}, \textit{wife}, \textit{husband}, \textit{grandchild}, \textit{granddaughter}, \textit{grandson}, \textit{aunt}, \textit{uncle}, \textit{niece}, \textit{nephew}, \textit{cousin}). Then, we added more entities, starting from a list of job titles,\footnote{The job titles originate from  \url{https://github.com/jneidel/job-titles/tree/master}, which collects job titles from a variety of sources.} which are all naturally human nouns. We then used word frequency data from 1 million words of Wikipedia \citep{goldhahn-etal-2012-building}, filtering out any job titles that are not found in that text. We then manually added valid job titles to our pool in order of descending frequency, until our expanded pool numbered 100 entries: 21 generic human entities and 79 professions. We could have increased the pool size by including more jobs, but the job title pool was noisy, and needed manual filtering to weed out nonsensical jobs, and jobs that sound too much like inanimate objects.

\begin{figure}
    \centering
    \includegraphics[width=\linewidth]{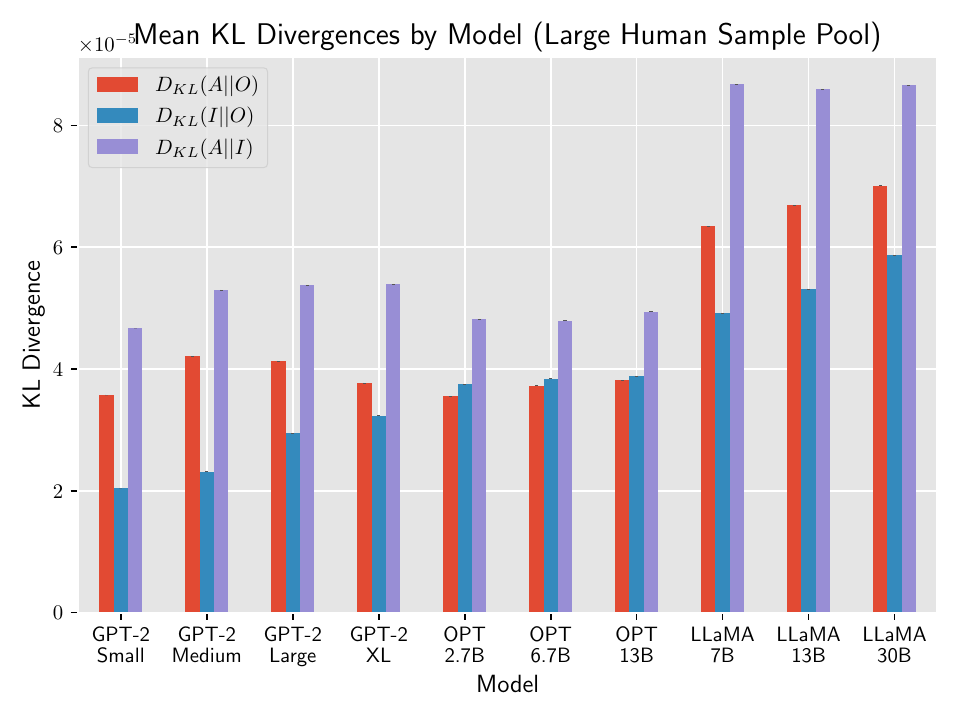}
    \caption{KL Divergence between atypically animate ($O$) and animate ($A$) / inanimate ($I$) references, where the animate entity in $A$ is drawn from a larger pool. The lower the bar, the more similar the distributions.}
    \vspace{-5pt}
    \label{fig:exp3-bars-large-pool}
\end{figure}

Then, we conducted our experiment again as in \Cref{sec:low-context}, using this larger pool. The results of this experiment (\Cref{fig:exp3-bars-large-pool}) are rather similar to that of our original experiment. In all cases the output distribution $p_O$ is clearly more animate than $p_I$; in the OPT models, the $p_O$ is more like the animate ($p_A$, red) than inanimate ($p_I$, blue) distribution. However, the animacy divergence is somewhat higher for the LLaMA models.

\subsection{Matched Frequency Human Entity Experiment}
In this experiment, we ensure that differences between the frequency of the sampled human entity and that of the inanimate entity do not undermine our experimental setup. Using the same frequency data as before, we matched each inanimate entity in our pool to the (manually verified, valid) human entity with the most similar frequency. We excluded one inanimate object (``well'') because its frequency was confounded by the very common adverb that shares its form. We were able to find a good human match for each inanimate object: the object frequency-to-human frequency ratio ranged from 0.92 to 1.09; this is near 1, the ideal ratio.

\begin{figure}
    \centering
    \includegraphics[width=\linewidth]{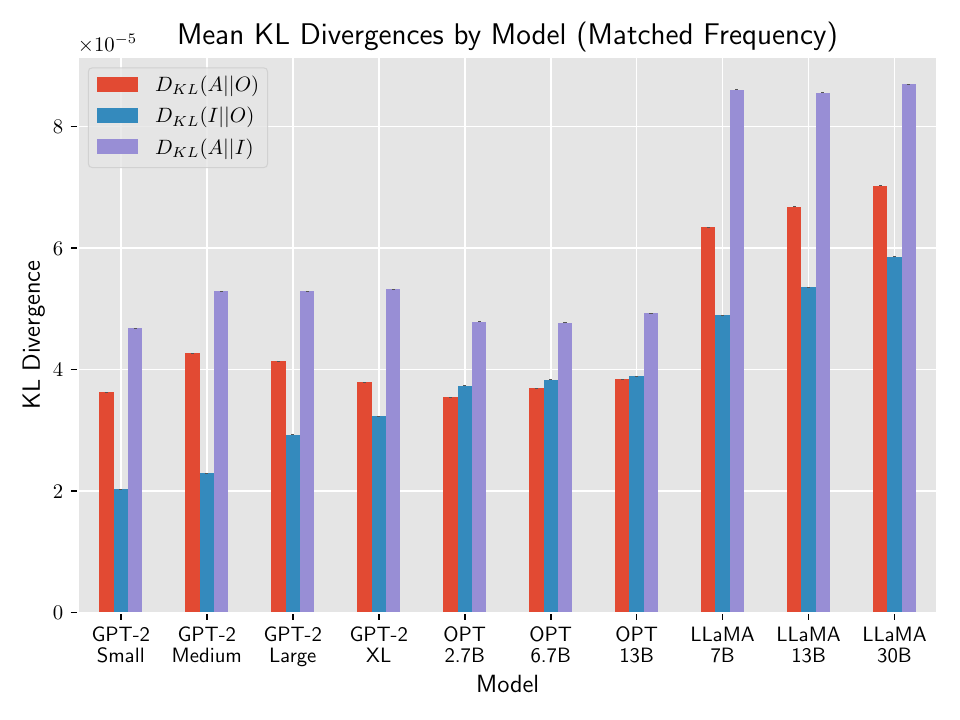}
    \caption{KL Divergence between atypically animate ($O$) and animate ($A$) / inanimate ($I$) references, where the animate entity in $A$ is matched in frequency with the inanimate entity in $I$ and $O$. The lower the bar, the more similar the distributions.}
    \vspace{-5pt}
    \label{fig:exp3-bars-matched}
\end{figure}

Then, we conducted our experiment again as in \Cref{sec:low-context}, using this larger pool. The results of this experiment (\Cref{fig:exp3-bars-matched}) are strikingly similar to those of the previous experiment (\Cref{fig:exp3-bars-large-pool}). We take this to suggest that frequencies are indeed not very important to the phenomenon we observe.

\subsection{Cataphoric Prompt Experiment}
In our original experiment, we test sentences of the form $O=$``The chair spoke and began to''. One potential concern is that LMs might just look at ``spoke and began to'', which implies an animate continuation, thus overcoming the inanimacy of \emph{chair}. We can test this by using a cataphoric prompt, where a referring pronoun comes before the object, e.g. $O'=$``After it spoke, the chair began to''. In such a prompt, there now exists the $4$-gram ``the chair began to'', which should make this task a little harder for LMs. Using the less-animate / inanimate pronoun ``it'' to refer to the atypically animate object also makes this more challenging.

\begin{figure}
    \centering
    \includegraphics[width=\linewidth]{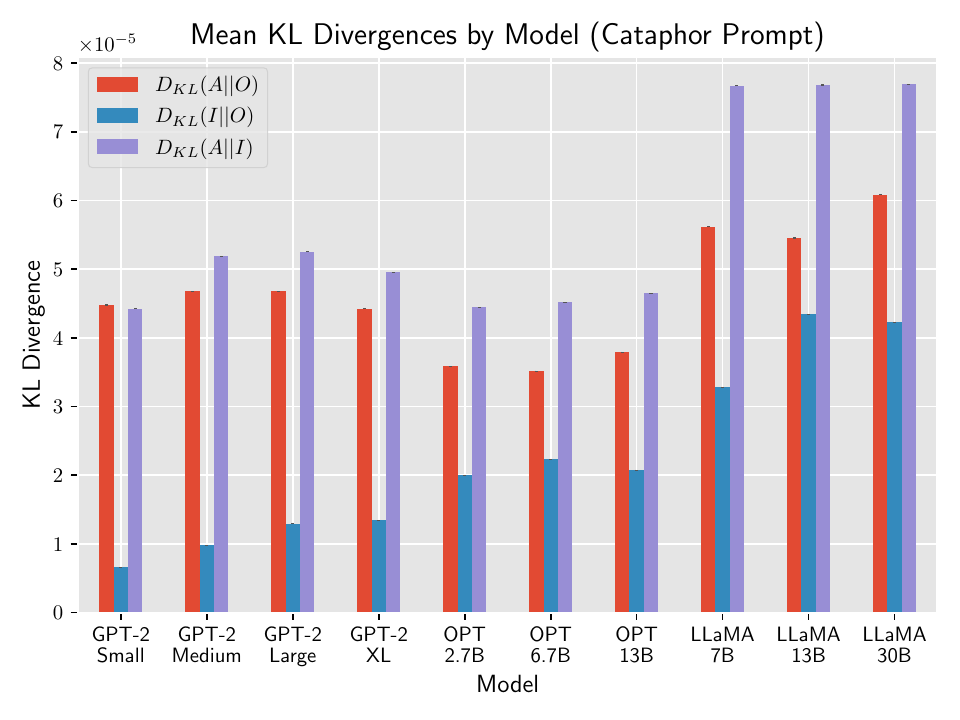}
    \caption{KL Divergence between atypically animate ($O$) and animate ($A$) / inanimate ($I$) references, using the cataphoric prompt. The lower the bar, the more similar the distributions. }
    \vspace{-5pt}
    \label{fig:exp3-bars-cataphor}
\end{figure}

Then, we conducted our experiment again as in \Cref{sec:low-context}, using the original small pool of human entities. The results of this experiment (\Cref{fig:exp3-bars-cataphor}) suggest that this setting is indeed harder for LMs; no longer to any LMs adapt such that divergence between $p_O$ and $p_A$ (red) is less than that between $p_O$ and $p_I$ (blue). However, there is still an increase in animacy compared to the case where the prompt does not hint at atypical animacy (purple).

\section{Low-context Animacy: Evaluation}\label{app:low-context-evaluation}
We first tried to classify potential next tokens as animate, inanimate, or neither; we could then compute the probability assigned to each group. However, classifying all next tokens was noisy: even with resources like FrameNet or VerbNet \citep{baker-etal-1998-berkeley-framenet,kipper2000verbnet}, it was infeasible to determine if a verb implied that its subject was animate.

\section{Low-Context Animacy Dataset Details}\label{app:low-context-dataset}
This section contains lists of the prompts, nouns, and verbs used in constructing this dataset. The full dataset, and all variants thereof, can be found at \url{https://github.com/hannamw/lms-in-love}.

\subsection{Prompts} 
\begin{itemize}
    \item The \texttt{[noun]} \texttt{[verb]} and began to
    \item The \texttt{[noun]} \texttt{[verb]} and started to
    \item The \texttt{[noun]} \texttt{[verb]} and was very
    \item The \texttt{[noun]} \texttt{[verb]} and became very
\end{itemize}
\subsection{Nouns} accordion, ambulance, amplifier, appliance, arrow, automobile, axe, bagpipe, balloon, bandage, banner, barrel, basket, bin, biscuit, blanket, blossom, blouse, boat, bomb, book, bottle, bouquet, bra, bracelet, bread, brush, bubble, bucket, buckle, bullet, button, cake, camera, candle, candy, cane, cannon, canoe, cape, cart, casket, chisel, chocolate, clarinet, clock, clothing, coat, cocktail, coffin, coin, collar, corpse, dagger, dart, desk, dime, dress, engine, envelope, ferry, fiddle, firewood, flask, flute, football, fruit, furniture, glass, glove, goblet, gown, hailstone, hairpin, hammer, harp, hat, helmet, hose, jar, keg, kilt, knife, lamp, lantern, lens, limousine, mallet, map, mattress, medallion, microscope, mirror, missile, moccasin, nail, napkin, necklace, needle, nickel, nightgown, oar, ornament, oven, overcoat, pants, pearl, pencil, pendulum, penny, phone, photograph, piano, pie, pillow, pipe, plank, pot, propeller, prune, purse, quilt, radio, record, refrigerator, ribbon, rifle, ring, rope, rug, sandal, satchel, saxophone, scissors, scroll, shawl, shield, shirt, shoe, ski, skull, sleigh, sock, sofa, spoon, statue, steak, stove, submarine, sword, tablespoon, telephone, telescope, thermometer, thorn, thread, ticket, tie, timepiece, tractor, tray, tripod, trombone, truck, trumpet, tube, tweezers, twig, typewriter, umbrella, van, vase, vehicle, vest, violin, wallet, wheel, whistle, wig, yacht, zipper, 
\subsection{Verbs}
\subsubsection{Physical}
\paragraph{High Co-Occurrence With Human Subjects} stammer, grimace, mumble, drawl, frown, gasp, yell, nod, smile, laugh, shrug, sob, grin, kneel, wince, whisper, sigh, giggle, squint, murmur, doze, fiddle, gesture, mutter, faint, gulp, flinch, chuckle, drink, weep, stare, grunt, listen, watch, fumble, shiver, pace, lean, blush, shout, gaze, walk, sit, sleep, dine, pant, glare, clap, stumble, snore, shave, wave, omit, sniff, piss, cough, wail, grumble, breathe, snort, spit, eat, duck, die, swallow, growl, blink, inhale, bellow, starve, crouch, yawn, step, squat
\paragraph{High-Mid Co-Occurrence With Human Subjects} pounce, scream, flee, shudder, wander, shriek, stagger, wink, sing, whistle, jog, limp, hiss, trot, jump, bathe, dance, paint, ramble, shower, drown, recover, pack, sweat, bow, flush, crawl
\paragraph{Mid Co-Occurrence With Human Subjects} bark, swim, bleed, howl 
\subsubsection{Psychological}
\paragraph{High Co-Occurrence With Human Subjects} think, know, wonder, remember, guess, exclaim, retort, marry, notice, understand, hurry, pray, meditate, swear, forget, enquire, realise, confess, apologise, hesitate, suspect, reply, talk, sneer, cry, dream, moan, ponder, revel, learn, scowl, retire, snarl, groan, speak, complain, beg, wait, preach, grieve, read, plead, volunteer, answer, curse, choose, panic, chant, cheat, salute, emigrate, protest, visit, lament, misunderstand
\paragraph{High-Mid Co-Occurrence With Human Subjects} consent, graduate, disagree, steal, mourn, study, argue, search, insist, practise, interrupt, obey, comment, concede, fight, applaud, enlist, worry, teach, train, agree, struggle, rush, evacuate, object, pay, pursue, hasten
\paragraph{Mid Co-Occurrence With Human Subjects} vote, invest, register

\end{document}